\documentclass[10pt,twocolumn,letterpaper]{article}

\usepackage[pagenumbers]{cvpr} 

\definecolor{cxcolor}{RGB}{  0, 204, 153}
\definecolor{xtcolor}{RGB}{255, 140,   0}

\newcommand{\sysname}[0]{EnvGS}
\newcommand{\envgs}[0]{\textit{environment Gaussian}}
\newcommand{\basegs}[0]{\textit{base Gaussian}}

\definecolor{cvprblue}{rgb}{0.21,0.49,0.74}
\definecolor{tabfirst}{rgb}{1, 0.7, 0.7}
\definecolor{tabsecond}{rgb}{1, 0.85, 0.7}

\definecolor{tabthird}{rgb}{1, 1, 0.7}
\newcommand{\cellfirst}[1]{\cellcolor{tabfirst}{#1}}
\newcommand{\cellsecond}[1]{\cellcolor{tabsecond}{#1}}
\newcommand{\cellthird}[1]{\cellcolor{tabthird}{#1}}

\usepackage[pagebackref,breaklinks,colorlinks,allcolors=cvprblue]{hyperref}

\usepackage[english]{babel}
\usepackage{colortbl}
\usepackage{xcolor} 
\usepackage{tabularx}
\usepackage{multirow}
\usepackage{booktabs}
\usepackage{graphicx}
\usepackage{array}
\usepackage{caption}
\usepackage{amsmath}
\usepackage{float}
\usepackage{booktabs}
\usepackage{capt-of}
\usepackage[accsupp]{axessibility}  

\title{EnvGS: Modeling View-Dependent Appearance with Environment Gaussian}

\author{
    Tao Xie$^{1*}$ \quad
    Xi Chen$^{1*}$ \quad
    Zhen Xu$^{1}$ \quad
    Yiman Xie$^{1}$ \quad
    Yudong Jin$^{1}$ \quad \\
    Yujun Shen$^{2}$ \quad
    Sida Peng$^{1}$ \quad
    Hujun Bao$^{1}$ \quad
    Xiaowei Zhou$^{1 \dagger}$
    \vspace{0.2cm}
\\
    {\normalsize $^{1}$ Zhejiang University} \quad
    {\normalsize $^{2}$ Ant Group} \quad
}

\begin{document}

\twocolumn[
    \maketitle
    \begin{center}
    \captionsetup{type=figure}
    \includegraphics[width=1.0\textwidth]{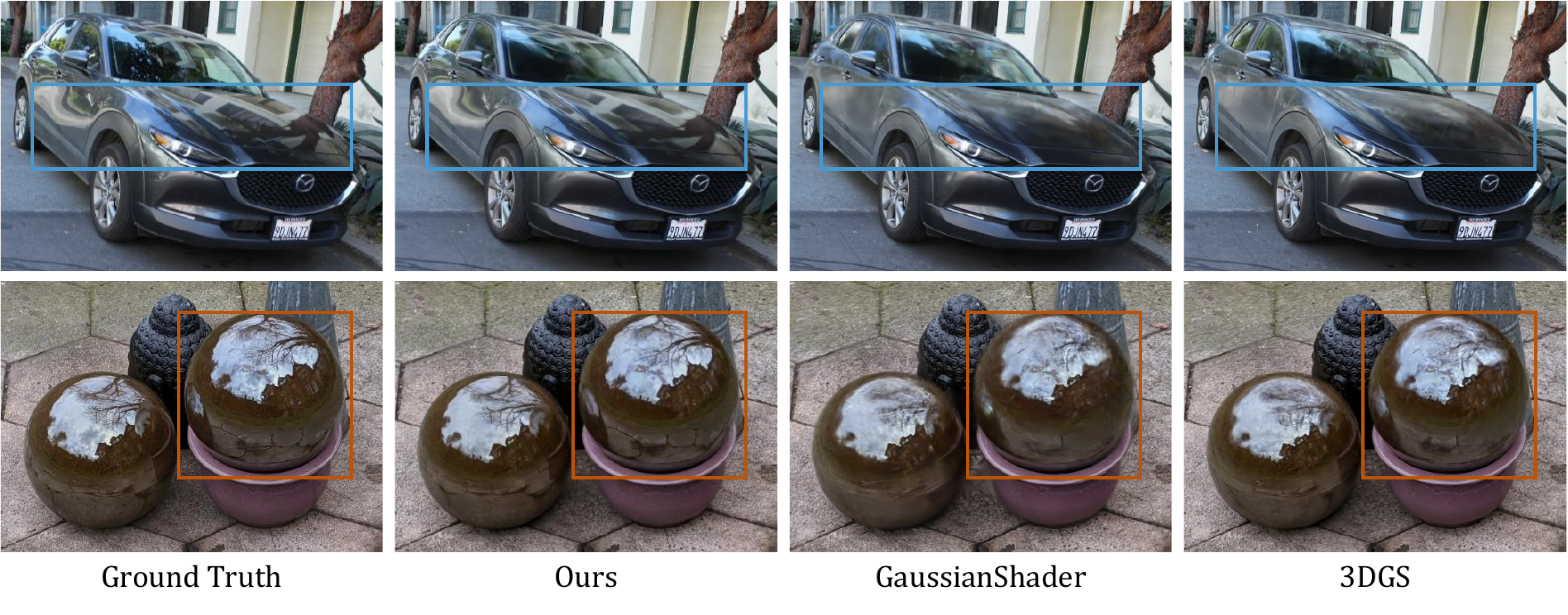}
    \caption{\textbf{Photorealistic, real-time rendering of real-world scenes with complex reflections.} Our proposed \sysname{} outperforms prior works in capturing complex reflection effects, especially near-field reflections and high-frequency details while maintaining real-time rendering speed. Please see our supplementary video for better visualizations.}
    \label{fig:teaser}
\end{center}

    \bigbreak
]

\newcommand\blfootnote[1]{%
  \begingroup
  \renewcommand\thefootnote{}\footnote{#1}%
  \addtocounter{footnote}{-1}%
  \endgroup
}
\blfootnote{\textcolor{black}{*} Equal Contribution. \textcolor{black}{$\dagger$} Corresponding author: Xiaowei Zhou}

\begin{abstract}
Reconstructing complex reflections in real-world scenes from 2D images is essential for achieving photorealistic novel view synthesis. 
Existing methods that utilize environment maps to model reflections from distant lighting often struggle with high-frequency reflection details and fail to account for near-field reflections. 
In this work, we introduce \sysname, a novel approach that employs a set of Gaussian primitives as an explicit 3D representation for capturing reflections of environments.
These environment Gaussian primitives are incorporated with base Gaussian primitives to model the appearance of the whole scene.
To efficiently render these environment Gaussian primitives, we developed a ray-tracing-based renderer that leverages the GPU's RT core for fast rendering. 
This allows us to jointly optimize our model for high-quality reconstruction while maintaining real-time rendering speeds. 
Results from multiple real-world and synthetic datasets demonstrate that our method produces significantly more detailed reflections, achieving the best rendering quality in real-time novel view synthesis.
The code is available at \href{https://zju3dv.github.io/envgs}{https://zju3dv.github.io/envgs}.
\end{abstract}
    
\section{Introduction}
\label{contents:introduction}

Novel view synthesis aims to generate novel views of 3D scenes based on a set of input images, which enables many applications such as VR/AR, and autonomous driving.
Recent advances in Neural Radiance Fields (NeRF)~\cite{mildenhall2020nerf} have demonstrated impressive rendering performance. 
However, NeRF's high computational cost makes it challenging for real-time applications.
More recently, 3D Gaussian Splatting (3DGS)~\cite{kerbl3Dgaussians} explicitly models scenes with 3D Gaussian primitives and utilizes rasterization for rendering, achieving real-time rendering with competitive quality.
However, modeling complex high-frequency specular reflections remains challenging for 3DGS due to the limited expressiveness of the Spherical Harmonics (SH).

Recent works GaussianShader~\cite{jiang2024gaussianshader} and 3DGS-DR~\cite{ye20243d}, enhance 3DGS by integrating an environment map and employing shading functions to blend the appearance from both the environment map and SH for the final rendering. While additional environmental lighting can enhance 3DGS's reflection modeling ability, it still struggles to reconstruct complex specular reflections accurately due to two factors.
First, the assumption of distant lighting in environment maps limits their ability to only capture distant illumination and difficult to synthesize accurate near-field reflections. Second, this representation inherently lacks sufficient capacity to capture high-frequency reflection details.

In this paper, we present \sysname{}, a novel approach for modeling complex reflections in real-world scenes, addressing the aforementioned challenges.
We propose to model reflections using a set of Gaussian primitives termed the \envgs{}. The geometry and base appearance are represented by another set of Gaussian primitives called \basegs{}. We effectively blend the two Gaussians for rendering and optimization.
Our rendering process begins with rendering the \basegs{} for the per-pixel surface position, normal, base color, and blending weight. Next, we render the \envgs{} at the surface point in the direction of the reflection of the viewing direction around the surface normal to capture reflection colors. Finally, we blend the base color with the reflection color to achieve the final rendering results.
In contrast to previous methods, \sysname{} captures high-frequency reflection details using Gaussian primitives, offering superior modeling capabilities compared to environment maps. Additionally, our explicit 3D reflection representation eliminates the need for distant lighting assumptions, enabling accurate modeling of near-field reflections, as shown in Fig.~\ref{fig:teaser}. 

To render the \envgs{} at each intersection point along the reflection direction, we create a fully differentiable ray-tracing renderer for 2DGS since rasterization is not suited for this task.
We build the ray-tracing renderer on CUDA and OptiX~\cite{parker2010optix} for real-time rendering and efficient optimization of \envgs{}.
The rendering process starts by constructing a bounding volume hierarchy (BVH) from the 2D Gaussian primitives. We then cast rays against the BVH, gathering ordered intersections in chunks while integrating the Gaussian properties through volume rendering~\cite{kajiya1984ray} to achieve the final results.
Our Gaussian ray-tracing renderer enables detailed reflection rendering at real-time performance. Furthermore, it allows for efficient joint optimization of the \envgs{} and \basegs{}, which is essential for accurate reflection modeling, as demonstrated in Sec~\ref{exp:ablation}.

To validate the effectiveness of our method, we evaluate \sysname{} on several real and synthetic datasets. The results demonstrate that our methods achieve state-of-the-art performance in real-time novel view synthesis and considerably surpass existing real-time methods, particularly in synthesizing complex reflections in real-world scenes.

In summary, we make the following contributions:
\begin{itemize}
    \item We propose a novel scene representation for accurately modeling complex near-field and high-frequency reflections in real-world environments.
    \item We developed a real-time ray-tracing renderer for 2DGS, enabling joint optimization of our representation for accurate scene reconstruction while achieving real-time rendering speeds.
    \item Extensive experiments shows that \sysname{} significantly outperforms previous methods. To the best of our knowledge, \sysname{} is the first method that achieves real-time photorealistic specular reflections synthesizing in real-world scenes.
\end{itemize}

\section{Related Work}

In computer vision and graphics research~\cite{Ye_Liu_Shen_2024,Hang_Rui_Shipeng_Hao_Heng_2023,Xing_Xu_2024,Hu_Li_Zhou_Cheng_Gu_2025,Li_Hu_Xu_Xu_Wang_2024,Wang_Zuo_Zhao_Lyu_Liu_2022,lin2024promptda}, our work falls into the area of learning scene representations from a set of posed RGB images. 
In this section, we review NeRF-based and Gaussian Splatting-based methods, particularly their handling of view-dependent effects.
\label{contents/related_work}

\paragraph{Neural Radiance Field.} NeRF~\cite{mildenhall2020nerf} introduced the concept of neural radiance fields, which model scenes as implicit multilayer perceptrons (MLPs) and render them via volume rendering, achieving impressive results in novel view synthesis.
Subsequent advancements focused on enhancing rendering quality ~\cite{barron2021mipnerf, barron2022mipnerf360,barron2023zipnerf}  and computational efficiency. ~\cite{mueller2022instant, Chen2022ECCV, yu2022plenoxels, SunSC22, liu2020neural}.
\cite{yariv2020multiview, wang2021neus, neus2, li2023neuralangelo} introduce Signed Distance Field into NeRF to improve geometry quality.
However, these methods often model view-dependent effects via simple viewing directions, which can lead to blurry reflection renderings.
To address this, Ref-NeRF~\cite{verbin2022refnerf} encodes the outgoing radiance using the reflected view direction, yielding improved results under distant lighting conditions.
Follow-up works~\cite{liu2023nero, wang2023unisdf, liang2023envidr, gcchen2024pisr} leverage Signed Distance Fields (SDF) to refine surface normals to enhance reflection 
 and geometry quality. 
SpecNeRF~\cite{ma2023specnerf} further incorporates spatially varying Gaussian directional encoding to better capture near-field reflections.
NeRF-Casting~\cite{verbin2024nerf} represents the scene in a unified manner, similar to~\cite{barron2023zipnerf}, and performs ray marching along reflection directions to integrate reflection features, which are then decoded into color with MLPs.
This approach achieves impressive results on real-world data under both near and distant lighting conditions.
However, the requirement for multiple MLP queries per ray makes NeRF-Casting unsuitable for real-time rendering and necessitates substantial training time
In contrast, our approach delivers real-time rendering capabilities while significantly reducing training time.

\paragraph{Gaussian Splatting.} Recently, 3D Gaussian Splatting (3DGS)~\cite{kerbl3Dgaussians} has made significant strides toward real-time novel view synthesis. 
Unlike volume-based rendering, Gaussian Splatting uses efficient rasterization to render spatial Gaussian kernels, achieving real-time performance at 1080P resolution.
Mip-Splatting~\cite{Yu2023MipSplatting} further introduces a 3D smoothing filter and a 2D Mip filter for alias-free renderings. Scaffold-GS~\cite{scaffoldgs} introduces structured 3D Gaussians to improve rendering efficiency and quality. \cite{bi2024gs3} introduces triple splitting for efficient relighting.
2D Gaussian Splatting (2DGS) ~\cite{huang20242d} replaces the 3D Gaussian kernels with 2D Gaussian that better align with 3D surfaces, leading to more accurate surface reconstruction.
However, these methods face challenges in accurately modeling reflections due to their use of spherical harmonics (SH) to parameterize view-dependent effects based solely on viewing direction, which often results in blurry reflections.
More recent works GaussianShader~\cite{jiang2023gaussianshader} and 3DGS-DR~\cite{ye2024gsdr} try to incorporate additional environment maps to improve reflection modeling ability.
However, these methods only consider distant lighting, ignoring near-field lighting, and are unable to capture high-frequency reflection details.
3iGS~\cite{tang20243igs} extends 3DGS by augmenting it with an illumination field using tensorial factorization, rendering final reflections through a neural renderer. 
However, it is limited to bounded scenes, which restricts its applicability in real-world, unbounded environments.
Concurrent work~\cite{zhang2024ref} choose to model near-field reflection with tensorial factorization and far-field with spherical feature grid.
Our method adopts 2DGS as the scene representation and models environmental illumination using an additional set of environment Gaussian, which is rendered via our proposed Gaussian tracer.
Our approach inherently supports high-frequency, near-field, and distant lighting in unbounded scenes, enabling detailed reflection rendering while maintaining real-time performance.

\begin{figure*}[ht!]
  \centering
  \includegraphics[width=1.0\linewidth]{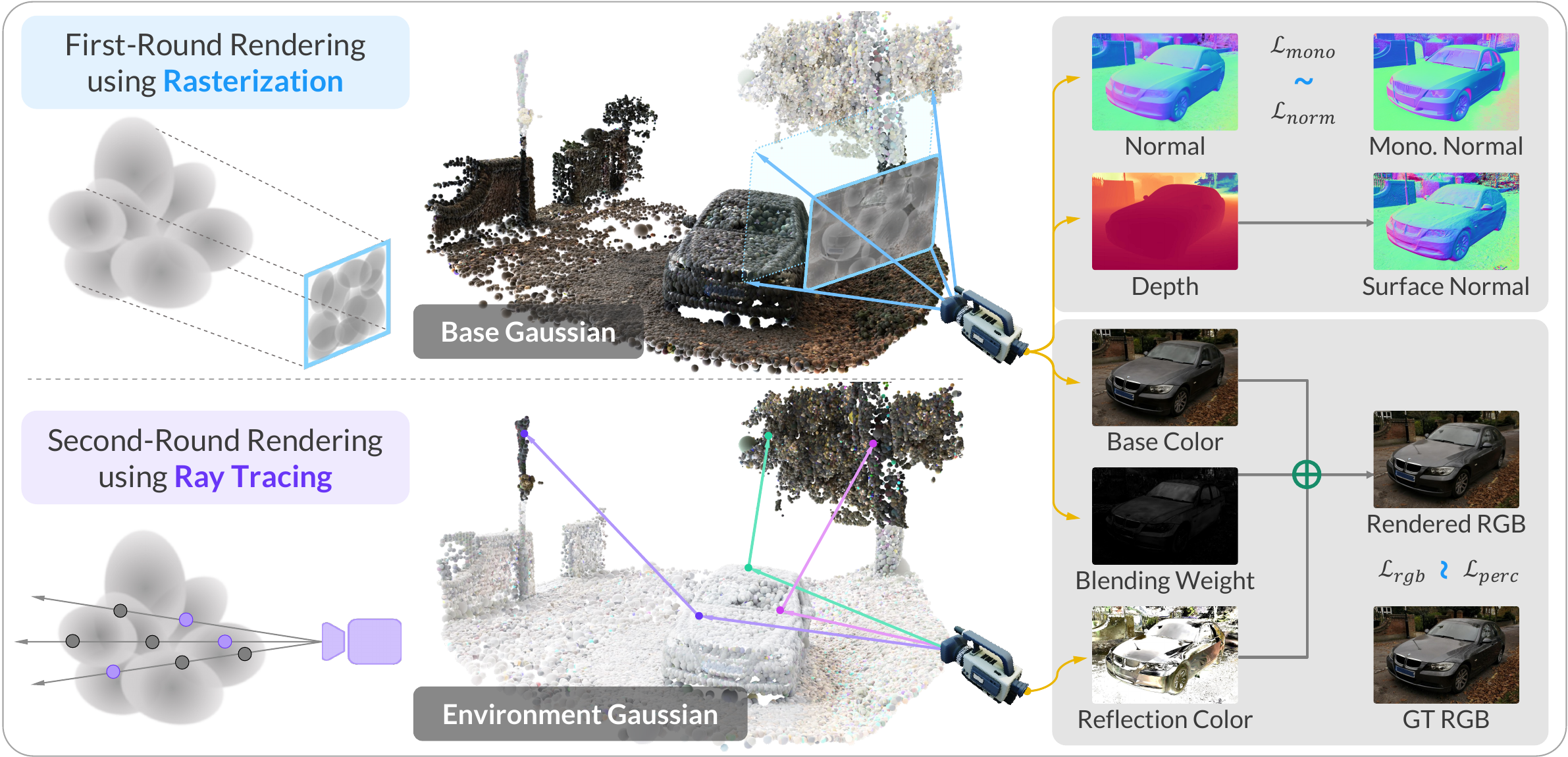}
  \caption{\textbf{Overview of \sysname{}}. The rendering process begins by rasterizing the \basegs{} to obtain per-pixel normals, base colors, and blending weights. Next, we render the \envgs{} in the reflection direction using our ray-tracing-based Gaussian renderer to capture the reflection colors. Finally, we combine the reflection and base colors for the final output. We jointly optimize the \envgs{} and \basegs{} using monocular normals~\cite{ye2024stablenormal} and ground truth images for supervision.}
  \label{fig:pipeline}
  \vspace{-10pt}
\end{figure*}

\section{Preliminary}
\label{contents:preliminary}

We begin by introducing 2D Gaussian Splatting (2DGS)~\cite{huang20242d}, which our approach is built upon.
2DGS is an explicit scene representation similar to 3D Gaussian Splatting (3DGS)~\cite{kerbl3Dgaussians}, which uses Gaussian primitives and rasterization to render screen projections.
The key difference is that 2DGS represents the scene with a set of scaled 2D Gaussian primitives defined in a local tangent plane within world space by a transformation matrix $\mathbf{H}$:
\begin{equation}
  \begin{aligned}
    \mathbf{H} &= \begin{bmatrix}
    s_{u} \mathbf{t}_{u} & s_{v} \mathbf{t}_{v} & 0 & \mathbf{p}_{k} \\
    0 & 0 & 0 & 1
    \end{bmatrix},
  \end{aligned}
  \label{eq:transformation}
\end{equation}
where $\mathbf{p}_k$, $(\mathbf{t}_u, \mathbf{t}_v)$, and $(s_u, s_v)$ denote the center, the two principal tangential vectors, and the scaling factors of the Gaussian, respectively.

To render an image, 2DGS employs the method described in~\cite{weyrich2007hardware} to determine the ray-primitive intersections. These intersection points are subsequently utilized to compute the Gaussian's contribution to the final image.
The Gaussian properties are then integrated using a volume rendering algorithm to obtain the final pixel color:
\begin{equation}
  \begin{aligned}
    \mathbf{c} = \sum_{i=1}^N T_i \alpha_i \mathbf{c}_i, \ \textrm{with} \ \alpha_i = \sigma_i \mathcal{G}_i, \ T_i = \prod_{j=1}^{i-1} (1 - \alpha_j),
  \end{aligned}
  \label{eq:volume_rendering}
\end{equation}
where $\mathcal{G}(\cdot)$ is the standard 2D Gaussian value evaluation.

Compared to 3D Gaussian, 2D Gaussian offers distinct advantages as a surface representation. First, the ray-splat intersection method adopted by 2DGS avoids multi-view depth inconsistency. Second, 2D Gaussian inherently provides a well-defined normal, which is essential for high-quality surface reconstruction and accurate reflection calculations.
However, 2DGS relies on the limited representational capacity of Spherical Harmonics (SH) to model scene appearance, preventing it from capturing strong view-dependent effects like specular reflections, which leads to poor rendering results and ``foggy" geometry~\cite{verbin2022refnerf}.
To this end, we use the geometry-aligned 2D Gaussian primitives as the base scene representation and demonstrate how we effectively model complex reflections in the next section.

\section{Method}
\label{contents:method}
Given a set of input images of a reflective scene, our goal is to reconstruct the 3D scene and synthesize photorealistic novel views in real-time.
To achieve this, we propose utilizing \envgs{} as an explicit 3D environment representation, which enables accurate modeling of complex reflections in real-world scenes. Additionally, we represent the scene geometry and base colors using another set of Gaussian primitives, denoted as \basegs.
In this section, we first detail how \envgs{} and \basegs{} work together to model complex reflections within the scene (Sec.~\ref{method:representation}). Then, We describe the design of a ray-tracing renderer that leverages the GPU's RT cores to efficiently render and optimize the \envgs{} (Sec.~\ref{method:tracing}).
Finally, we discuss our optimization process. (Sec.~\ref{method:optimization}).
An overview of our method is in Fig.~\ref{fig:pipeline}.

\subsection{Reflective Scenes Modeling}
\label{method:representation}

Gaussian splatting~\cite{kerbl3Dgaussians, huang20242d} models appearance using Spherical Harmonics (SH), which has limited representation capacity for view-dependent effects. These limitations hinder its ability to capture complex, high-frequency specular reflections.
Building on this observation, our key insight is that modeling reflections with a Gaussian environment representation can better model complex reflection effects while significantly reducing the complexity required for each Gaussian to capture intricate details within its SH.

Our proposed reflective scene representation includes two sets of 2D Gaussians: a \basegs{} $\mathbf{P}_{base}$ for modeling the scene's geometry and base appearance, and another \envgs{} $\mathbf{P}_{env}$ for capturing scene reflections.
The basic parameterization of each Gaussian primitive is consistent with the original 2DGS~\cite{huang20242d}, including 3D center position $\mathbf{p}$, opacity $\alpha$, two principal tangential vectors $(\mathbf{t}_u, \mathbf{t}_v)$, a scaling vector $(s_u, s_v)$, and SH coefficients.
To combine the two appearance components from the base and environment Gaussian into the final result, we introduce a blending weight for each base Gaussian.

\begin{figure}[ht!]
    \centering
    \includegraphics[width=1.0\linewidth]{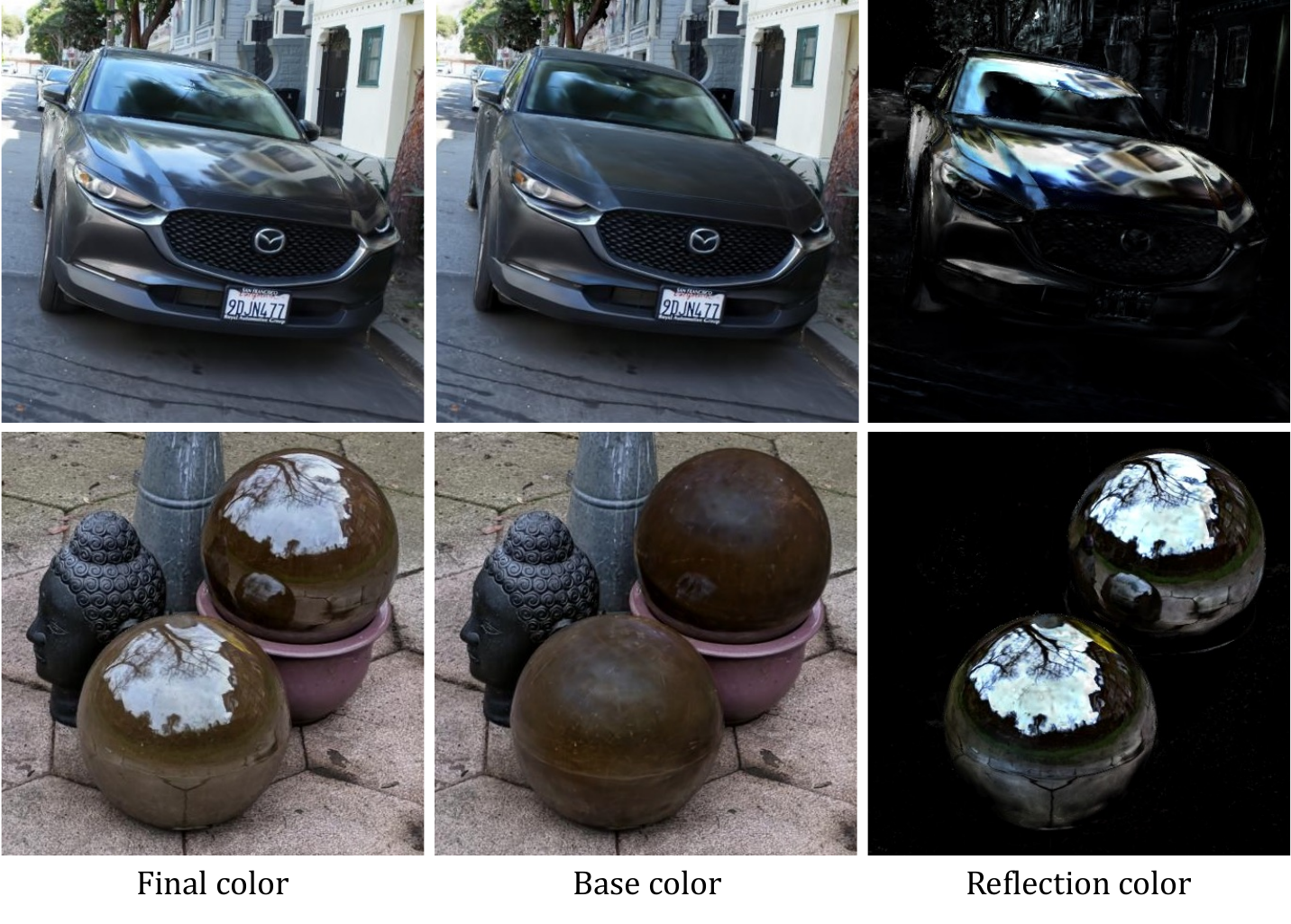}
    \caption{\textbf{Visualization of reflection and base color.} Our method successfully reconstructs near-field and distant reflections using the \envgs{} instead of baking into the base color.}
    \label{fig:decompose}
    \vspace{-9pt}
\end{figure}

The rendering process is performed in three steps. First, the base Gaussian $\mathbf{P}_{base}$ is rendered using standard 2D Gaussian splatting to obtain the base color $\mathbf{c}_{base}$. By applying the volume rendering integration using Eq. (\ref{eq:volume_rendering}), we also derive the surface position $\mathbf{x}$, surface normals $\mathbf{n}$, and blending weight $\beta$ as:
\begin{equation}
  v = \sum_{i \in \mathcal{N}} = v_{i} \alpha_{i} \prod_{j=1}^{i-1}\left(1-\alpha_{j}\right),\ v\in\{\mathbf{x}, \mathbf{n}, \beta\}.
\end{equation}
Then, we compute the reflection direction $\mathbf{d}_{ref}$ based on the camera ray direction $\mathbf{d}_{ref}$ and the surface normal $\mathbf{n}$:
\begin{equation}
  \mathbf{d}_{ref} = \mathbf{d}_{cam} - 2(\mathbf{d}_{cam} \cdot \mathbf{n}) \mathbf{n}.
\end{equation}
With the reflection direction $\mathbf{d}_{ref}$ and surface point $\mathbf{x}$, the environment Gaussian $\mathbf{P}_{env}$ is rendered using our differentiable Gaussian tracer to obtain the reflection color $\mathbf{c}_{ref}$, as detailed in Sec.~\ref{method:tracing}.
The final color is obtained through:
\begin{equation}
  \mathbf{c} = (1 - \beta) \cdot \mathbf{c}_{base} + \beta \cdot \mathbf{c}_{ref}.
  \label{eq:blendering}
\end{equation}
We blend the base color $\mathbf{c}_{base}$ and reflection color $\mathbf{c}_{ref}$ using the blending weight $\beta$. A visualization of $\mathbf{c}_{base}$ and $\mathbf{c}_{ref}$ can be found at Fig.~\ref{fig:decompose}.

\noindent\textbf{Discussion.} Compared to the environment map used by Gaussianshader~\cite{jiang2024gaussianshader} and 3DGS-DR~\cite{ye20243d}, our explicit Gaussian environment representation offers several advantages.
First, our method more accurately captures near-field reflections caused by occlusions from nearby objects. This improvement arises from explicitly modeling each Gaussian at its exact spatial location, thus avoiding the ambiguities and inaccuracies inherent in environment map representations that assume distant lighting.
Second, by utilizing Gaussian primitives, \sysname's environment representation achieves greater expressiveness than low-frequency environment maps, enabling the capture of finer reflection details and enhancing rendering quality, as demonstrated by our experiments~\ref{exp:baselines}.

\subsection{Differentiable Ray Tracing}
\label{method:tracing}
Rendering the environment Gaussian with rasterization is impractical, as each pixel corresponds to a unique reflection ray and functions as a virtual camera. To address this, we draw on~\cite{3dgrt2024} and leverage the advanced optimizations of modern GPUs to design a novel, fully differentiable ray tracing framework.
Built on OptiX~\cite{parker2010optix}, our framework achieves real-time rendering of 2,000,000 2DGS with a resolution of 1292x839 at 30 FPS on an RTX 4090 GPU.

In order to fully utilize the hardware acceleration for ray-primitive intersections, we need to convert each 2D Gaussian into a geometric primitive compatible with GPU processing and insert it into a bounding volume hierarchy (BVH).
In light of this, we propose to represent each 2D Gaussian with two triangles. Specifically, we first define the four Gaussian bounding vertices $\mathbf{V}_{local}=\{(\operatorname{sgn}(r), \operatorname{sgn}(r))\}$ in the local tangent plane, where $\operatorname{sgn}(\cdot)$ is the sign function and $r$ is set to 3 representing three times the sigma range. Then, the four local bounding vertices $\mathbf{V}_{local}$ are transformed to world space as the vertices of the two triangles covering the Gaussian $\mathbf{V}_{world}$ using Eq.~\ref{eq:transformation}.
After the transformation, the triangles are organized into a BVH, which serves as the input for the ray tracing process.

We develop a custom CUDA kernel using the \textit{raygen} and \textit{anyhit} programmable entry points of OptiX. Inspired by~\cite{3dgrt2024}, the rendering is done in a chunk-by-chunk manner.
The \textit{anyhit} kernel traces the input ray to obtain a chunk of size k, while \textit{raygen} integrates this chunk and invokes \textit{anyhit} to retrieve the next chunk along the ray.
Specifically, consider an input ray with origin $\mathbf{o}$ and direction $\mathbf{d}$.
The \textit{raygen} program first initiates a traversal against the BVH to identify all possible intersections along the ray. 
During the traversal, the \textit{anyhit} program sorts each intersected Gaussian by the depth and maintains a sorted k-buffer for the closest k intersections.
We empirically found that k with 16 is the best trade-off between traversal counts and the number of Gaussians sorted per traversal.
After traversal, the \textit{raygen} program integrates properties of the sorted Gaussians in the buffer following Eq.~\ref{eq:volume_rendering}. 
The Gaussian response is calculated by applying the inverse of the transformation matrix $\mathbf{H}$ to the ray's intersection point $\mathbf{x}_i$ and evaluating the Gaussian value at the transformed point as:
\begin{equation}
  \begin{aligned}
    \mathcal{G}_i(\mathbf{u}_i) = \mathcal{G}_i(\mathbf{H}^{-1}\mathbf{x}_i).
  \end{aligned}
  \label{eq:uv}
\end{equation}
This process repeats until no further intersections are found along the ray or the accumulated transmittance drops below a specified threshold. More details can be found in supplementary materials.

Our ray tracing framework is fully differentiable, allowing for end-to-end optimization of both the base and the environment Gaussian primitives.
However, storing all intersections during the forward pass and performing the backward pass in back-to-front order, as done in 3DGS~\cite{kerbl3Dgaussians}, is impractical due to high memory consumption.
To address this, we implement the backward pass in the same front-to-back order as the forward pass by re-casting rays and computing gradients for each integration step. A key aspect is calculating the gradient with respect to the input ray origin $\frac{\partial \mathcal{L}}{\partial \mathbf{o}}$ and direction $\frac{\partial \mathcal{L}}{\partial \mathbf{d}}$, which is crucial for the joint optimization of our model (see supplementary for more details). Our ablation in Sec~\ref{exp:ablation} demonstrates that this joint optimization of geometry and appearance is essential for recovering geometrically accurate surfaces.

\subsection{Optimization}
\label{method:optimization}
To enhance training stability, we initiate optimization by first training the base Gaussian $\mathbf{P}_{base}$, which is initialized using the sparse point cloud obtained from Structure-from-Motion (SfM)~\cite{snavely2006photo,he2024detector}.
After the bootstrapping phase, we initialize the environment Gaussian $\mathbf{P}_{env}$ by partitioning the scene's bounding box $\mathbf{B}_{scene}$ into $N^3$ sub-grids and randomly sampling $K$ primitives within each grid, and then optimize the base Gaussian and environment Gaussian jointly.
The scene's bounding box $\mathbf{B}_{scene}$ is determined by the $99.5\%$ quantile of the sparse point cloud's bounding box $\mathbf{B}_{sfm}$ obtained from SfM, the sub-grids resolution $N$ is set to 32, and $K$ is set to 5 for each grid.
\begin{figure*}[htbp]
    \centering
    \includegraphics[width=1.0\linewidth]{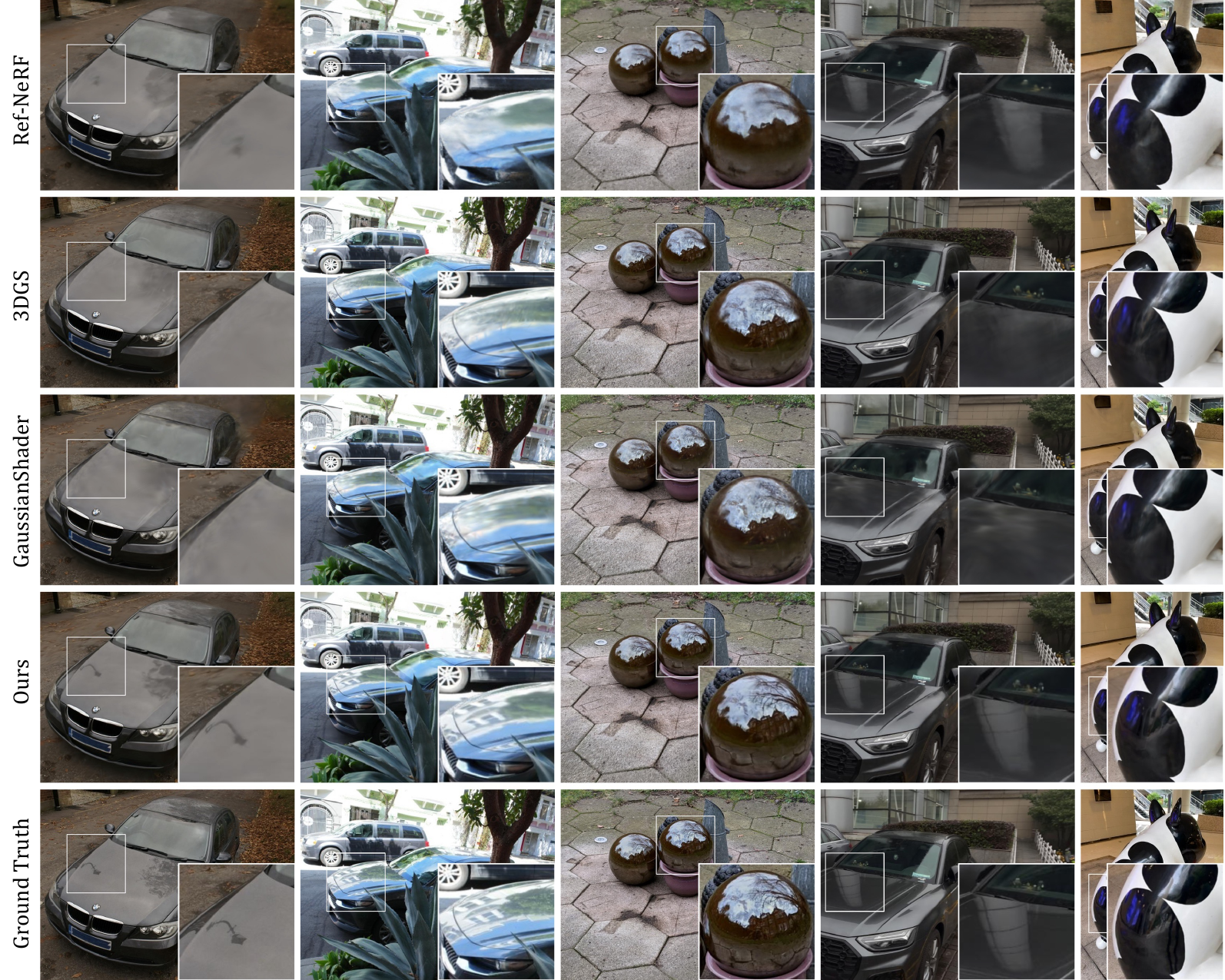}
    \caption{\textbf{Qualitative comparison on real scenes.} Our method significantly improves rendering quality over previous approaches, particularly in producing more detailed reflections. Zoom in for more details.}
    \label{fig:comparisons}
    \vspace{-12pt}
\end{figure*}

We follow 2DGS~\cite{huang20242d} to add a normal consistency constraint between the rendered normal map $\mathbf{n}$ and the gradients of the depth map $\mathbf{N}_d$:
\begin{equation}
  \mathcal{L}_{norm} = \frac{1}{N_p} \sum_{i=1}^{N_p} (1 - \mathbf{n}_i^\top \mathbf{N}_d),
\end{equation}
where $N_p$ is the number of pixels in the image and $\mathbf{N}_d$ is calculated as finite differences of neighboring pixels in the depth map:
\begin{equation}
  \mathbf{N}_d(\mathbf{u}) = \frac{\nabla_u \mathbf{p}_d \times \nabla_v \mathbf{p}_d}{\left\| \nabla_u \mathbf{p}_d \times \nabla_v \mathbf{p}_d \right\|}.
\end{equation}

However, the normal consistency constraint alone is insufficient for accurately modeling ambiguous surfaces involving both reflection and refraction. Inspired by recent advances in monocular normal estimation~\cite{bae2024dsine,ye2024stablenormal,he2024lotus}, we propose to supervise the rendered normal map $\mathbf{n}$ using monocular normal estimates $\mathbf{N}_{m}$:
\begin{equation}
  \mathcal{L}_{mono} = \frac{1}{N_p} \sum_{i=1}^{N_p} (1 - \mathbf{n}_i^\top \mathbf{N}_m).
\end{equation}

In addition to the $\mathcal{L}_1$ image loss and D-SSIM loss $\mathcal{L}{ssim}$ employed by 3DGS~\cite{kerbl3Dgaussians}, we also use perceptual loss~\cite{zhang2018unreasonable} to enhance the perceived quality of the rendered image:
\begin{equation}
  \mathcal{L}_{perc}=\|\Phi(\mathbf{I})-\Phi(\mathbf{I}_{gt})\|_1,
\end{equation}
where $\Phi$ is the pre-trained VGG-16 network~\cite{simonyan2014very} and $\mathbf{I}, \mathbf{I}_{gt}$ are the rendered and ground truth images, respectively.

The final loss function is defined as:
\begin{equation}
    \mathcal{L} = \mathcal{L}_{rgb} + \lambda_{1} \mathcal{L}_{norm} + \lambda_{2} \mathcal{L}_{mono} + \lambda_{3} \mathcal{L}_{perc},
\end{equation}
where $\mathcal{L}_{rgb}$ is the photometric reconstruction loss combining the $\mathcal{L}_{1}$ loss and a D-SSIM~\cite{wang2004image} term with a ratio of $0.8$ and $0.2$ respectively. We set $\lambda_{1}=0.04$ , $\lambda_{2}=0.01$, and $\lambda_{3}=0.01$ across our experiments.

\section{Experiments}
\label{contents:experiments}

\begin{table*}[t]
  \setlength{\tabcolsep}{4pt} 
  \centering 
  \footnotesize  

      \begin{tabular}{l l ccc | ccc | c | c}
          \toprule
          & \multirow{2.5}{*}{\bfseries Methods} & \multicolumn{3}{c|}{\bfseries Ref-NeRF Real Scenes~\cite{verbin2022refnerf}} & \multicolumn{3}{c|}{\bfseries NeRF-Casting Scenes~\cite{verbin2024nerf}} & \multirow{2.5}{*}{\bfseries FPS $\uparrow$} & \multirow{2.5}{*}{\bfseries Training Time $\downarrow$} \\

          \cmidrule(lr){3-5} \cmidrule(lr){6-8}
          & & PSNR$\uparrow$ & SSIM $\uparrow$ & LPIPS $\downarrow$ & PSNR $\uparrow$ & SSIM $\uparrow$ & LPIPS $\downarrow$ & \\

          \midrule
          \multirow{4}{*}{\textit{Non real-time}}
          &Ref-NeRF*~\cite{verbin2022refnerf} &23.087           &\cellthird0.625           &\cellthird0.261           &30.583           &\cellsecond0.890           &\cellsecond0.124           &$\textless$0.1 &47h \\
          &UniSDF~\cite{wang2023unisdf}       &\cellsecond23.700           &\cellsecond0.635           &0.266           &\cellthird30.838           &\cellthird0.889           &0.130           &$\textless$0.1 &$\textgreater$47h \\
          &ZipNeRF~\cite{barron2023zipnerf}   &\cellthird23.677           &\cellsecond0.635           &\cellsecond0.247           &\cellfirst31.740 &\cellfirst0.904 &\cellfirst0.105 &$\textless$0.1 &$\textgreater$47h \\
          &NeRF-Casting~\cite{verbin2024nerf} &\cellfirst24.670 &\cellfirst0.659 &\cellfirst0.246 &\cellsecond31.023           &\cellthird0.889           &\cellthird0.128           &$\textless$0.1 &$\textgreater$47h \\

          \midrule
          \multirow{5}{*}{\textit{Real-time}}
          &3DGS~\cite{kerbl3Dgaussians}                  &\cellthird23.700           &\cellthird0.641           &\cellsecond0.262           &\cellsecond28.860           &\cellsecond0.877           &\cellsecond0.159           &\textbf{182.307} &\textbf{0.6h} \\
          &2DGS~\cite{huang20242d}                       &\cellsecond23.804           &\cellsecond0.654           &0.281           &28.276           &\cellthird0.862           &\cellthird0.193           &159.188          &0.7h \\
          &GaussianShader~\cite{jiang2023gaussianshader} &22.875           &0.622           &0.314           &26.412           &0.835           &0.216           &27.945           &1.6h \\
          &3DGS-DR~\cite{ye20243d}                       &23.522           &0.640           &\cellthird0.274           &\cellthird28.487           &0.858           &0.197           &133.593          &1.0h \\
          &Ours                                          &\cellfirst24.617 &\cellfirst0.671 &\cellfirst0.241 &\cellfirst30.444 &\cellfirst0.886 &\cellfirst0.148 &26.221           &2.5h \\
          \bottomrule
      \end{tabular}
  \caption{
    \textbf{Quantitative comparison on Ref-Real~\cite{verbin2022refnerf} and  NeRF-Casting Scenes~\cite{verbin2024nerf} datasets.} Our method delivers the highest rendering quality among real-time techniques and outperforms several non-real-time methods, achieving competitive results with the state-of-the-art non-real-time method NeRF-Casting~\cite{verbin2024nerf}, while being 100 times faster.
    Note that Ref-NeRF* is an improved version of Ref-NeRF~\cite{verbin2022refnerf} that uses Zip-NeRF's geometry model.
    All metrics are evaluated at 1/4 resolution as prior works ~\cite{verbin2024nerf}.
  }
  \label{tab:refreal_nerfcasting}

  \vspace{-8pt} 
\end{table*}

\subsection{Implementation Details}
\label{exp:implementation}
We implement \sysname{} with custom OptiX kernels and optimize our model using the PyTorch framework~\cite{paszke2019pytorch,xu2023easyvolcap} with the Adam optimizer~\cite{kingma2014adam}. Specifically, we set the learning rates for the parameters of each base and environment Gaussian to match those used in 2DGS~\cite{huang20242d}. The learning rate for the blending weight is set to $1e^{-2}$. During training, we apply the adaptive Gaussian control strategy of 3DGS~\cite{kerbl3Dgaussians} with the normal propagation and color sabotage introduced in 3DGS-DR~\cite{ye2024gsdr}. Since our Gaussian tracer integrates the Gaussian properties directly in 3D space, there is no valid gradient for the projected 2D center, which is used as the densification criterion in 3DGS. Following \cite{3dgrt2024}, we accumulate the 3D spatial gradients of the Gaussian position to achieve a similar effect. Note that each accumulated gradient is scaled by half of the intersection depth to prevent under-densification in distant regions. All experiments are conducted on a single NVIDIA RTX 4090 GPU.

\begin{table}[t]
  \setlength{\tabcolsep}{3pt}
  \centering  
  \footnotesize

  \resizebox{0.9\columnwidth}{!}{  
      \begin{tabular}{@{}l ccc | ccc}  
          \toprule
          \multirow{2.5}{*}{Methods} &\multicolumn{3}{c|}{\textbf{Foreground}} &\multicolumn{3}{c}{\textbf{Near-Field}} \\
          \cmidrule(lr){2-4} \cmidrule(lr){5-7}
          &PSNR$\uparrow$ &SSIM $\uparrow$ &LPIPS $\downarrow$ &PSNR $\uparrow$ &SSIM$\uparrow$ &LPIPS$\downarrow$ \\

          \midrule
          3DGS~\cite{kerbl3Dgaussians} &\cellthird31.681 &\cellsecond0.955 &\cellsecond0.046 &\cellsecond44.161 &\cellsecond0.994  &\cellsecond0.008 \\
          2DGS~\cite{huang20242d} &31.328 &\cellthird0.953 &0.053 &43.624 &\cellsecond0.994 &0.010 \\
          GaussianShader~\cite{jiang2023gaussianshader} &30.694 &0.947 &0.058 &42.391 &\cellthird0.993 &0.010 \\
          3DGS-DR &\cellsecond31.814 &\cellthird0.953 &\cellthird0.050 &\cellthird44.007 &\cellsecond0.994 &\cellthird0.009 \\

          \midrule
          Ours &\cellfirst33.295 &\cellfirst0.962 &\cellfirst0.040 &\cellfirst46.392 &\cellfirst0.996 &\cellfirst0.007 \\
          \bottomrule
      \end{tabular}
  }

  \caption{\textbf{Quantitative results of foreground and near-field region on Ref-Real~\cite{verbin2022refnerf} and  NeRF-Casting Scenes~\cite{verbin2024nerf}.} See supplementary \cref{sup:mask_comparison} for more details and qualitative results.}
  \label{tab:mask_metric}
  \vspace{-15pt}  
\end{table}

\subsection{Datasets and Evaluation Metrics}
\label{exp:datasets}
We train and evaluate \sysname{} on a range of datasets with a focus on real-world scenes characterized by complex view-dependent effects. We evaluate the Ref-Real~\cite{verbin2022refnerf} and NeRF-Casting Shiny Scenes~\cite{verbin2024nerf} to demonstrate our method's ability for complex specular reflections in real-world scenes. We additionally captured two more real-world scenes for a more comprehensive evaluation.
We demonstrate that our methods can reconstruct detailed reflections on complex real-world scenes with real-time rendering speed.
Additionally, we evaluate our method on the synthetic Shiny Blender dataset~\cite{verbin2022refnerf}, which is rendered using environment maps. More evaluation results can be found in the supplementary materials.

We maintain consistent training and testing splits and image resolution across all datasets, following prior works~\cite{verbin2022refnerf,barron2022mipnerf360,kerbl3Dgaussians}. 
We use three commonly used metrics for evaluation: PSNR, SSIM~\cite{wang2004image}, and LPIPS~\cite{zhang2018unreasonable}.

\subsection{Baseline Comparisons}
\label{exp:baselines}
We compare our method with both implicit and explicit prior works, including Zip-NeRF~\cite{huang20242d}, 3DGS~\cite{kerbl3Dgaussians}, 2DGS~\cite{huang20242d}, which are designed for general scenes with primarily diffuse appearances, as well as with methods specifically tailored for scenes with strong specular reflections, including Ref-NeRF~\cite{verbin2022refnerf}, GaussianShader~\cite{jiang2023gaussianshader}, 3DGS-DR~\cite{ye2024gsdr}, and NeRF-Casting~\cite{verbin2024nerf}. We also compare with ENVIDR~\cite{liang2023envidr}, NDE~\cite{wu2024neural} these two object-level methods.


\begin{figure*}[!ht]
    \centering
    \includegraphics[width=\linewidth]{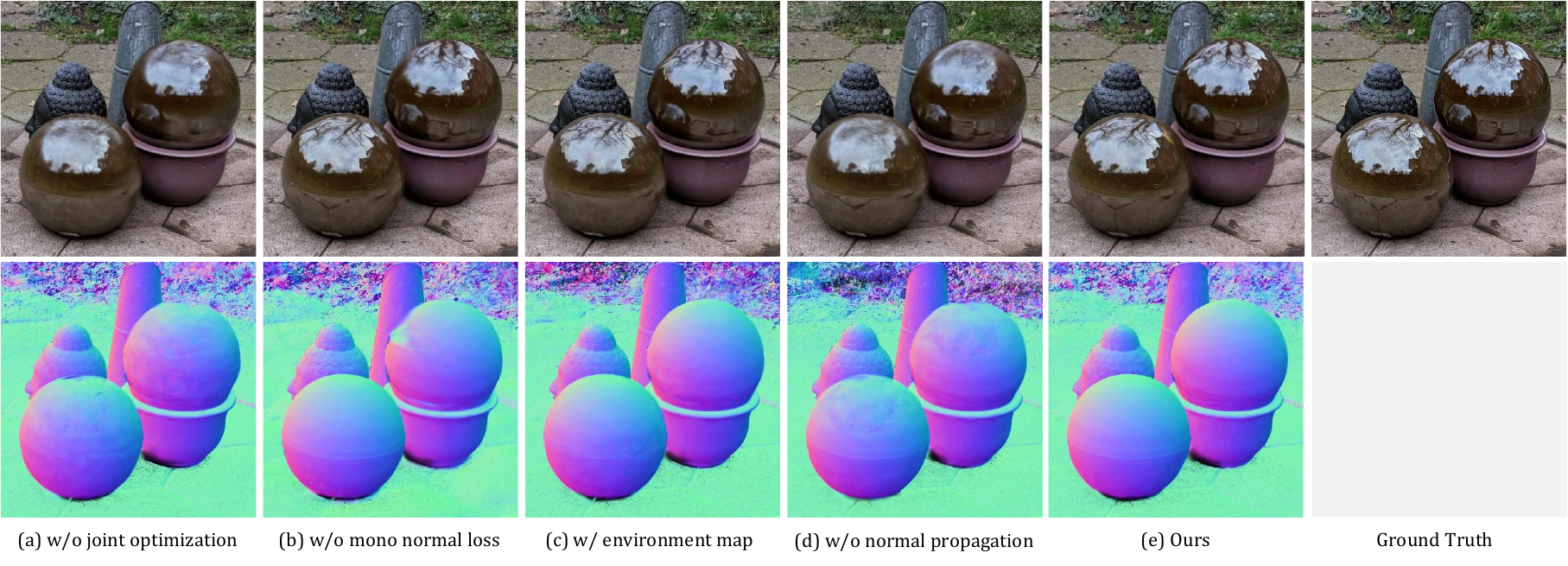}
    \vspace{-15pt}
    \caption{\textbf{Ablation study of proposed components on the Ref-Real dataset~\cite{verbin2022refnerf}}. Removing either the monocular normal constraint or the joint optimization of base and environment Gaussians results in noisy geometry and inaccurate reflection reconstruction. The ``\textit{w/} environment map'' variant fails to capture near-field reflections.}
    \label{fig:ablation_comparison}
    \vspace{-10pt}
\end{figure*}

In this section, we present quantitative and qualitative results to illustrate the advantages of our methods. Since accurate reflection reconstruction and rendering are the essential advantages of our method, we encourage readers to refer to the various rendered continuous video results in the supplementary material for a more comprehensive evaluation of our method.

\begin{table}[t]
  \renewcommand{\arraystretch}{1.0}
  \setlength{\tabcolsep}{3pt}
  \centering  
  \footnotesize

  \resizebox{0.9\columnwidth}{!}{  
      \begin{tabular}{@{}l ccc | ccc}  
          \toprule
          \multirow{2.5}{*}{Methods} &\multicolumn{3}{c|}{\textbf{Audi}} &\multicolumn{3}{c}{\textbf{Dog}} \\
          \cmidrule(lr){2-4} \cmidrule(lr){5-7}
          &PSNR$\uparrow$                    &SSIM $\uparrow$ &LPIPS $\downarrow$ &PSNR $\uparrow$ &SSIM$\uparrow$ &LPIPS$\downarrow$ \\

          \midrule
          Ref-NeRF~\cite{verbin2022refnerf}             &24.529 &0.713            &0.397            &23.620            &0.673            &0.389 \\
          3DGS~\cite{kerbl3Dgaussians}                  &\cellsecond26.171            &\cellsecond0.825 &\cellsecond0.181 &\cellsecond25.300 &\cellsecond0.882  &\cellsecond0.165 \\
          2DGS~\cite{huang20242d}                       &25.869            &0.819            &0.198            &\cellthird24.933            &\cellthird0.876            &\cellthird0.184 \\
          GaussianShader~\cite{jiang2023gaussianshader} &24.768            &0.795            &0.225            &23.061            &0.840            &0.226 \\
          3DGS-DR~\cite{ye20243d}                       &\cellthird26.093            &\cellthird0.820            &\cellthird0.192            &23.943            &0.850            &0.210 \\

          \midrule
          Ours                                          &\cellfirst28.629  &\cellfirst0.876  &\cellfirst0.121  &\cellfirst25.599  &\cellfirst0.885 &\cellfirst0.151 \\
          \bottomrule
      \end{tabular}
  }

  \caption{\textbf{Quantitative results on our self-captured scenes.}}
  \label{tab:our_scenes}
  \vspace{-15pt}  
\end{table}

We first evaluate our method on nine real-world scenes from the Ref-Real dataset~\cite{verbin2022refnerf}, NeRF-Casting Shiny Scenes~\cite{verbin2024nerf} and our self-captured real-world scenes, which feature complex geometry and specular reflections. The quantitative results, shown in Tab.~\ref{tab:refreal_nerfcasting} and Tab.~\ref{tab:our_scenes}, demonstrate that our method outperforms all explicit methods by a large margin and achieves comparable results to the current state-of-the-art implicit method, NeRF-Casting, while being significantly faster. The qualitative results in Fig.\ref{fig:comparisons} further highlight the superior performance of our method in capturing complex view-dependent effects, especially near-field reflections and high-frequency details.

We provide additional comparisons on the Mip-NeRF 360 dataset~\cite{barron2022mipnerf360}, the Shiny Blender dataset~\cite{verbin2022refnerf} and with ENVIDR~\cite{liang2023envidr}, NDE~\cite{wu2024neural} in supplementary \cref{sup:additional_results}, which also includes per-scene metric breakdowns from Tab.~\ref{tab:refreal_nerfcasting}. Our method achieves competitive results on both datasets, demonstrating its versatility and effectiveness in handling diverse scenes with complex view-dependent effects.

\subsection{Ablation Studies}

\begin{table}[h]
  \renewcommand{\arraystretch}{0.4}  
  \vspace{-5pt}  
  \centering  
  \footnotesize  

  \begin{tabular}{p{3cm}ccc}
      \toprule
      & PSNR $\uparrow$ & SSIM $\uparrow$ & LPIPS $\downarrow$ \\

      \midrule
      \textit{w/o} joint optimization &24.034           &0.644          &0.287            \\
      \textit{w/o} mono normal loss   &24.107           &0.648          &0.270            \\
      \textit{w/} environment map     &24.145           &0.646          &0.285            \\
      \textit{w/o} color sabotage     &24.268           &0.650          &0.269            \\
      \textit{w/o} normal propagation &24.192           &0.652          &0.271            \\
      \textit{w/o} lpips loss         &24.567           &\textbf{0.671} &0.280            \\
      \midrule
      Ours                            &\textbf{24.617}  &\textbf{0.671} &\textbf{0.241}   \\
      \bottomrule
  \end{tabular}
  \caption{\textbf{Ablation studies.}}
  \label{tab:ablation}

  \vspace{-10pt}  
\end{table}

\label{exp:ablation}
In this section, we conduct ablation studies of our key components on the Ref-Real dataset. Quantitative and qualitative results are shown in Tab.~\ref{tab:ablation} and Fig.~\ref{fig:ablation_comparison}, respectively.

\noindent\textbf{Joint optimization.} The ``\textit{w/o} joint optimization'' variant detaches the joint optimization of the \basegs{} and \envgs{} from the reflection rendering step. As shown in Tab.~\ref{tab:ablation} and Fig.~\ref{fig:ablation_comparison}, This variant fails to recover accurate geometry, leading to inferior reflection reconstruction and rendering quality.

\noindent\textbf{Monocular normal constraint} The ``\textit{w/o} monocular normal'' variant removes the monocular normal constraint as described in Sec.~\ref{method:optimization}. Training may become trapped in largely incorrect geometry, resulting in inaccurate reflection reconstruction, as demonstrated in Tab.\ref{tab:ablation} and Fig.\ref{fig:ablation_comparison}.

\noindent\textbf{Environment map representation} The ``\textit{w/} environment map'' variant replaces our core Gaussian environment representation with an environment map representation while keeping all other components unchanged. Figure~\ref{fig:ablation_comparison} illustrates that while it effectively captures smooth distant reflections, it has difficulty modeling near-field and high-frequency reflections, and produces more bumpy geometry.

\noindent\textbf{Color sabotage and normal propagation.} The ``\textit{w/o} color sabotage'' and ``\textit{w/o} normal propagation'' variants omit the color sabotage and normal propagation steps from our method, respectively. As demonstrated in Tab.~\ref{tab:ablation}, both variants result in reduced rendering quality.

\noindent\textbf{Perceptual loss.} The ``\textit{w/o} lpips loss'' variants removes the perceptual loss. The results in Tab.~\ref{tab:ablation} demonstrate that our method still produces detailed and accurate reflections without perceptual loss, which offers only marginal improvements.

\section{Conclusion and Discussion}
\label{contents:conclusion}

This paper introduced \sysname{}, a novel reflective scene representation for high-quality complex reflection capturing and real-time rendering. Our method explicitly models reflections with a set of \envgs{} primitives. 
The environment Gaussian primitives are used together with a set of base Gaussian primitives that model basic scene properties (geometry, base color, and blending weight) to model the appearance of the whole scene.
Furthermore, we develop a differentiable Gaussian ray tracer utilizing GPU's RT core to effectively optimize and render the \envgs{}. The proposed method demonstrates superior performance in capturing complex reflections across various datasets.

A limitation of \sysname{} is its difficulty with transparent and refractive materials, as it only addresses reflection direction. Future work could explore extending our method to accommodate these materials.

\vspace{1em}
\noindent\textbf{Acknowledgements.}
This work was partially supported by NSFC (No. U24B20154, 62402427), Ant Group, and Information Technology Center and State Key Lab of CAD\&CG, Zhejiang University.

{
    \small
    \bibliographystyle{ieeenat_fullname}
    \bibliography{main}
}

\clearpage
\setcounter{page}{1}
\maketitlesupplementary
\appendix

In the supplementary material, we provide more qualitative and quantitative results and per-scene breakdowns to demonstrate the effectiveness and robustness of our method (Sec.~\ref{sup:additional_results}). We also provide additional ablation studies to further analyze the key components of our method (Sec.~\ref{sup:additional_ablations}). Furthermore, we provide details on the gradient computation of our Gaussian tracer (Sec.~\ref{sup:gradient}).

Accurate and smooth reflection reconstruction and rendering are key advantages of our method. We strongly encourage readers to view the rendered continuous videos in the supplementary material for a more comprehensive understanding of its performance.

\section{Additional Results}
\label{sup:additional_results}

\subsection{Comparison on Reflective Regions}
\label{sup:mask_comparison}
\begin{figure}[H]
    \centering
    \includegraphics[width=1.0\linewidth]{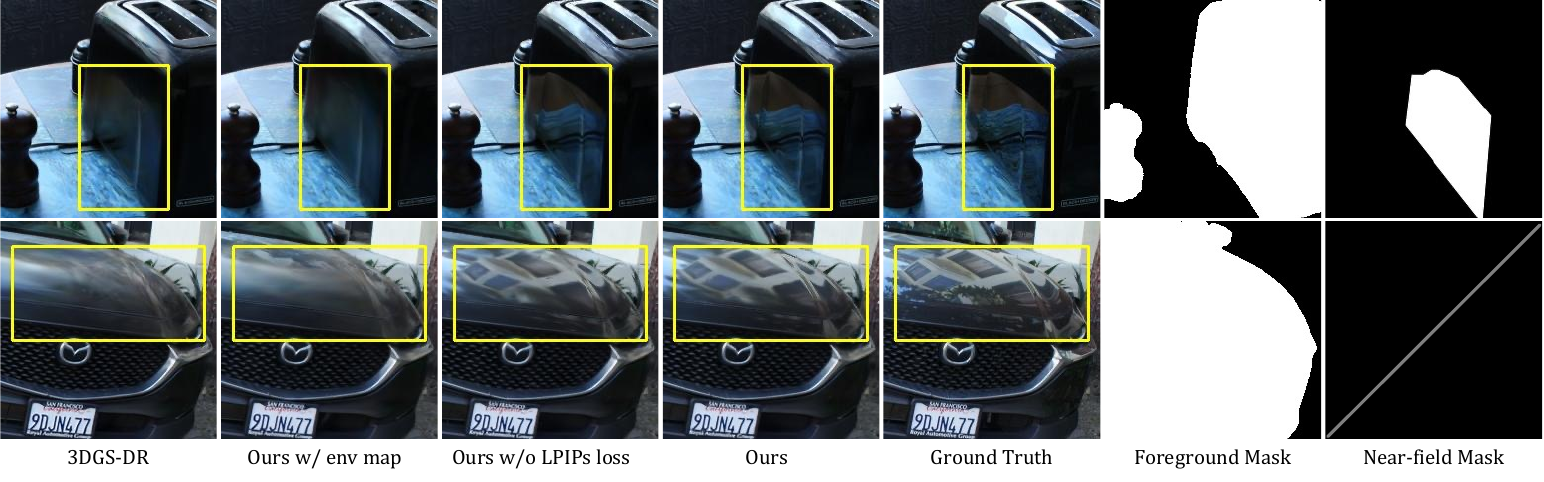}
    \caption{\textbf{Qualitative comparison on reflective foreground and near-field reflection regions.} We also provide visualizations of the foreground and near-field region mask we annotated.}
    \label{fig:reflective_region}
\end{figure}

To demonstrate the improvements in the reflective and near-field reflection regions using our environment Gaussian representation, we additionally annotate a reflection mask to compute metrics specifically for the reflective region and a near-field mask to evaluate near-field reflections on the Ref-Real~\cite{verbin2022refnerf} and NeRF-Casting~\cite{verbin2024nerf} datasets. As shown in \cref{tab:mask_metric} and \cref{fig:reflective_region}, our method achieves a significant improvement of over 1.0 PSNR $\uparrow$ improvement on the reflective region and 2.0 PSNR $\uparrow$ in the near-field regions compared to using an environment map. These results highlight the effectiveness of our approach in capturing and rendering complex reflective and near-field phenomena.

The reflective masks mentioned above are obtained through the following steps. First, we train our \sysname{} on each scene, then export the trained Gaussian and remove the Gaussian points in 3D space except for those in the foreground reflective region. We render the remaining Gaussian to generate an accumulated alpha map. Finally, we binarize this alpha map to obtain the foreground reflective masks. We manually annotate the near-field masks as they are difficult to define in 3D space.

The quantitative results in \cref{tab:mask_metric} are evaluated only in the masked regions, following NeRF-Casting~\cite{verbin2024nerf}, we compute these masked metrics by blending the masked regions onto a white background.

\subsection{Comparison on Real-World Shiny Scenes}
\label{sup:real_comparison}
\begin{figure*}[htbp]
    \centering
    \includegraphics[width=1.0\linewidth]{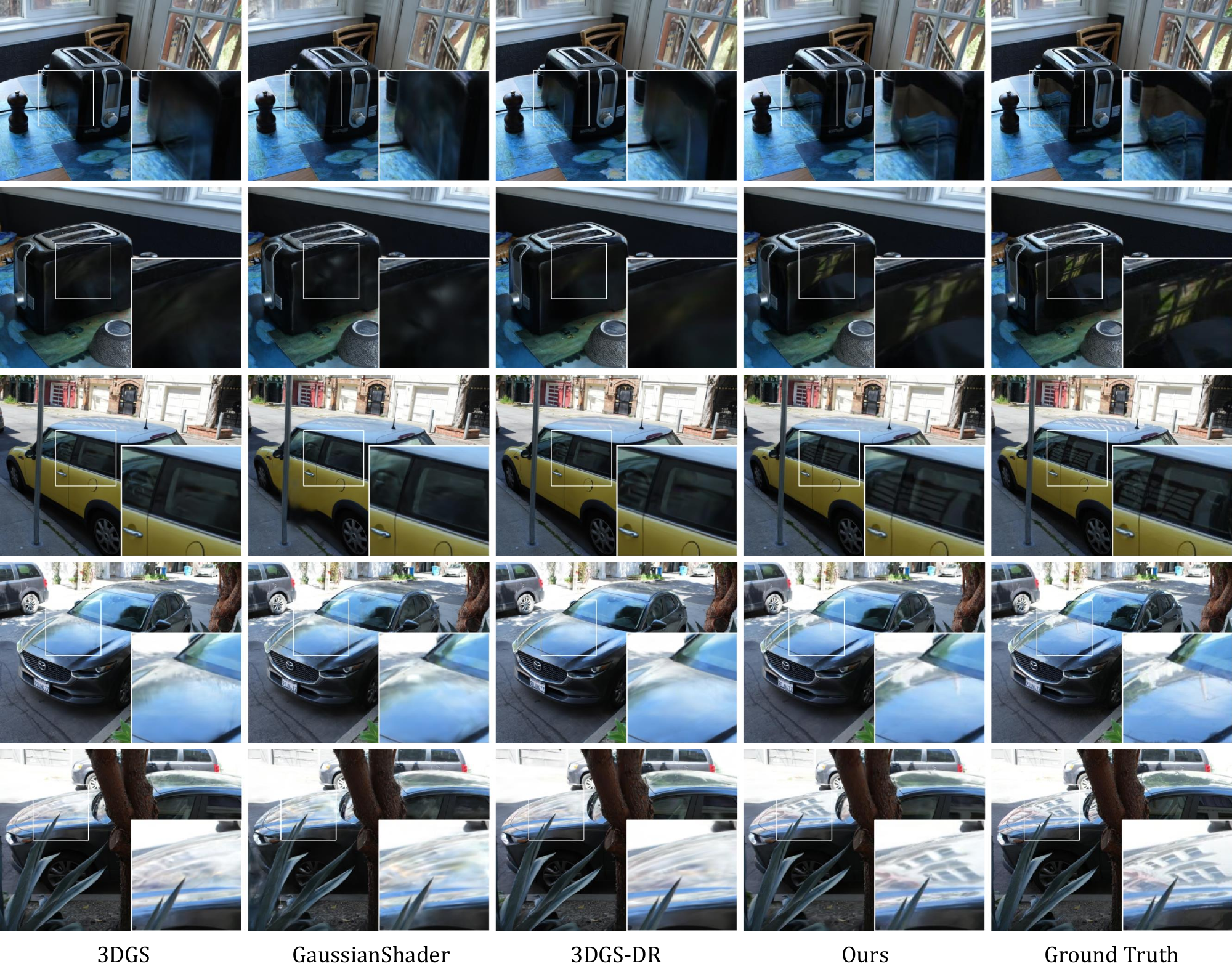}
    \caption{\textbf{Qualitative comparison on real scenes.} Our method significantly improves rendering quality over previous approaches, particularly in producing more detailed reflections. Zoom in for more details.}
    \label{fig:comparison_sup_real}
\end{figure*}

We present additional qualitative comparisons on the NeRF-Casting Shiny Scenes~\cite{verbin2024nerf}, including both indoor and outdoor real-world scenes featuring complex reflections. As shown in Fig.~\ref{fig:comparison_sup_real}, our method significantly outperforms previous approaches in reflection fidelity and overall rendering quality, particularly excelling in near-field reflections and high-frequency reflection details.

We also provide per-scene breakdowns of Ref-Real~\cite{verbin2022refnerf} and NeRF-Casting Shiny Scenes~\cite{verbin2024nerf} in Tab.~\ref{tab:real_casting_break}. These results are consistent with the averaged results in the paper. All metrics are evaluated at the original resolution downsampled by a factor of 4, following prior works~\cite{verbin2024nerf}. Notably, our method is more general and does not rely on manually estimated bounding boxes for foreground objects, which are essential for 3DGS-DR~\cite{ye20243d} to prevent optimization failure.

\subsection{Comparison on Shiny Blender~\cite{verbin2022refnerf}}
\label{sup:synthetic_comparison}
\begin{figure*}[h]
    \centering
    \includegraphics[width=1.0\linewidth]{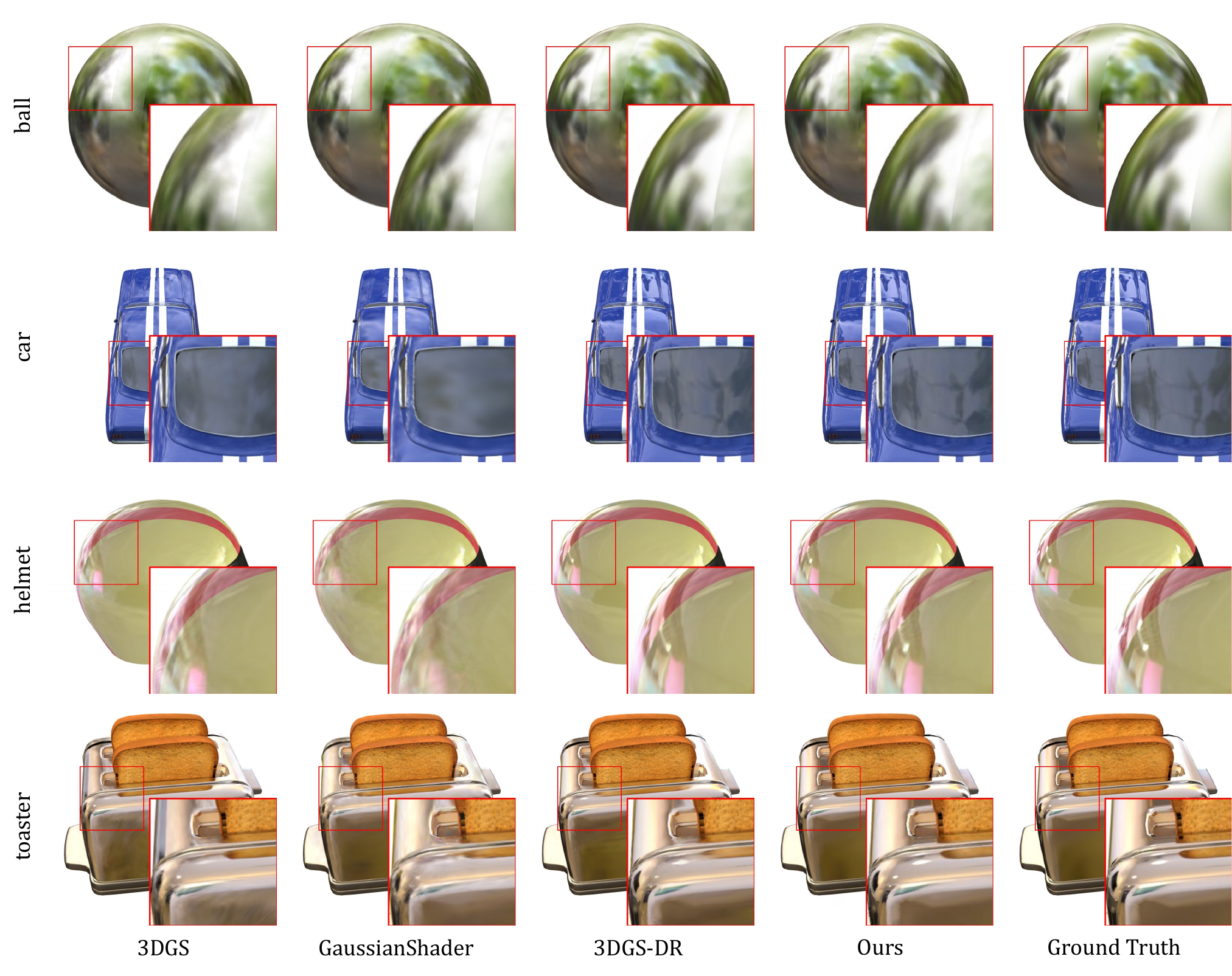}
    \caption{\textbf{Qualitative comparison on synthetic scenes.} Despite being designed for robustness on real-world data, our method effectively reconstructs accurate distant specular reflections and effectively captures near-field reflections caused by self-occlusions.}
    \label{fig:comparison_sup_synthetic}
\end{figure*}

\begin{figure*}[h]
    \centering
    \includegraphics[width=1.0\linewidth]{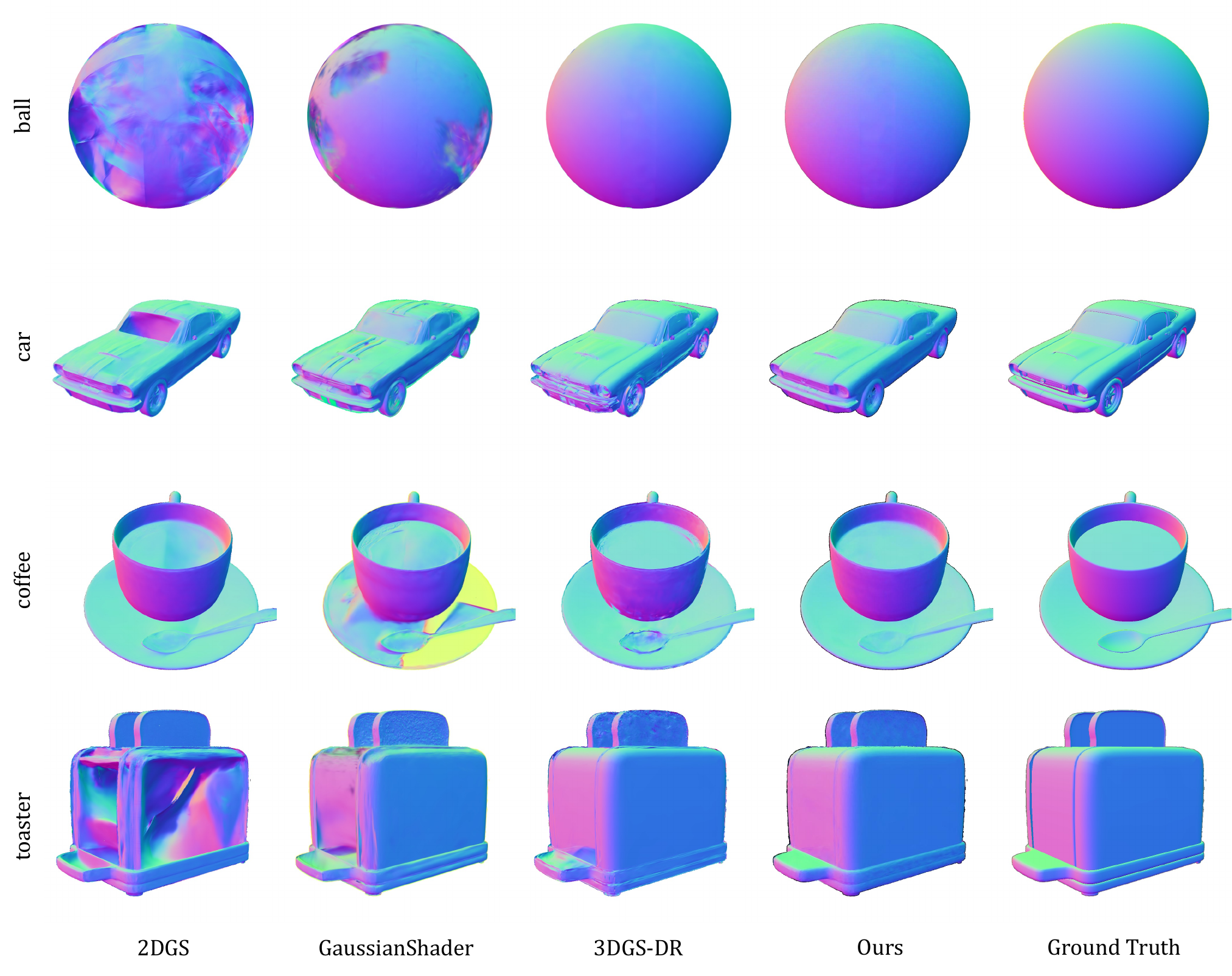}
    \caption{\textbf{Qualitative comparisons of normal produced by different methods.}}
    \label{fig:comparison_sup_synthetic_normal}
\end{figure*}

In Tab.~\ref{tab:shiny_blender}, Fig.~\ref{fig:comparison_sup_synthetic} and Fig.~\ref{fig:comparison_sup_synthetic_normal}, we present additional quantitative and qualitative comparisons on the Shiny Blender dataset~\cite{verbin2024nerf}, which is rendered with environment maps under distant lighting assumption. The results show that although being designed for robustness on real-world data, our method effectively reconstructs accurate distant specular reflections, performing on par with or surpassing prior methods GaussianShader~\cite{jiang2024gaussianshader} and 3DGS-DR~\cite{ye2024gsdr} specifically designed for environment map lighting scenarios. \sysname{} considerably outperforms these methods in capturing near-field reflections caused by self-occlusions, as illustrated in the zoomed-in regions of the ``toaster'' scene. Moreover, our method reconstructs more accurate geometry, as shown in Fig.~\ref{fig:comparison_sup_synthetic_normal}.

\subsection{Comparison on Mip-NeRF 360~\cite{barron2022mipnerf360}}
We perform additional comparisons on the Mip-NeRF 360 dataset~\cite{barron2022mipnerf360}, which consists of large-scale real-world scenes with primarily diffuse appearance and complex geometry. As shown in Tab.~\ref{tab:mipnerf360}, our method is not limited to reflective scenes and can achieve comparable or superior performance to both state-of-the-art implcit~\cite{barron2023zipnerf} and explicit~\cite{kerbl3Dgaussians} methods.

\begin{table}[t]
    \centering
    \footnotesize
    \renewcommand{\arraystretch}{0.8}

    \begin{tabular}{lcccc}
      \toprule
      \multirow{2.5}{*}{Methods} &\multicolumn{4}{c}{Ref-Real~\cite{verbin2022refnerf} and NeRF-Casting~\cite{verbin2024nerf}} \\
      \cmidrule{2-5} &PSNR $\uparrow$ &SSIM $\uparrow$ &LPIPS $\downarrow$ &FPS $\uparrow$ \\
      \midrule
      ENVIDR         &15.890           &0.416           &0.607           &0.058 \\
      NDE            &19.399           &0.422           &0.593           &0.083 \\
      Ours           &\textbf{27.947}  &\textbf{0.794}  &\textbf{0.189}  &\textbf{26.221} \\
      \bottomrule
    \end{tabular}

    \caption{\textbf{Quantitative comparison with object-level methods.}}
    \label{tab:object_baseline}
\end{table}

\begin{figure}[t]
  \centering
  \includegraphics[width=1.0\linewidth]{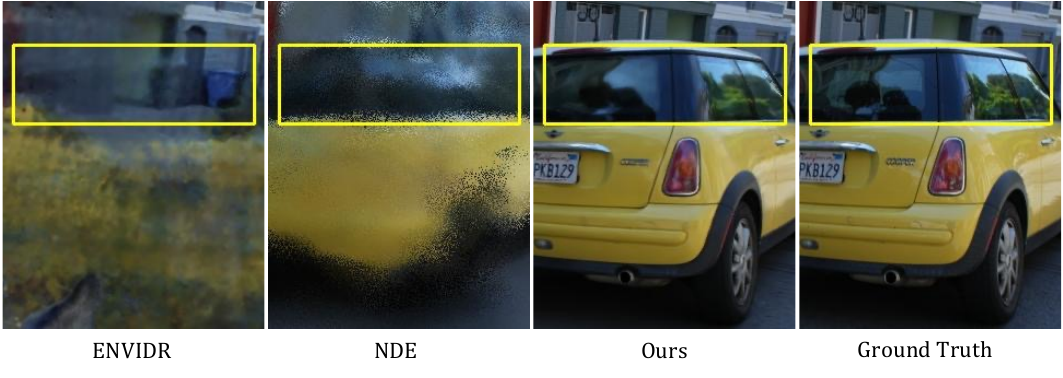}
  \caption{\textbf{Qualitative comparison with object-level methods.}}
  \label{fig:object_baseline}
\end{figure}

\subsection{Additional Baselines}
We also compare our method with object-baselines including ENVIDR~\cite{liang2023envidr} and NDE~\cite{wu2024neural}.
While object-level methods perform well on synthetic data, they often struggle with real-world scenes and cannot real-time rendering speed on scenes with background, as shown in \cref{tab:object_baseline} and \cref{fig:object_baseline}.

\section{Additional Ablation Studies}
\label{sup:additional_ablations}

\subsection{Environment Representation Comparison}


\begin{figure}[ht!]
    \centering
    \includegraphics[width=\linewidth]{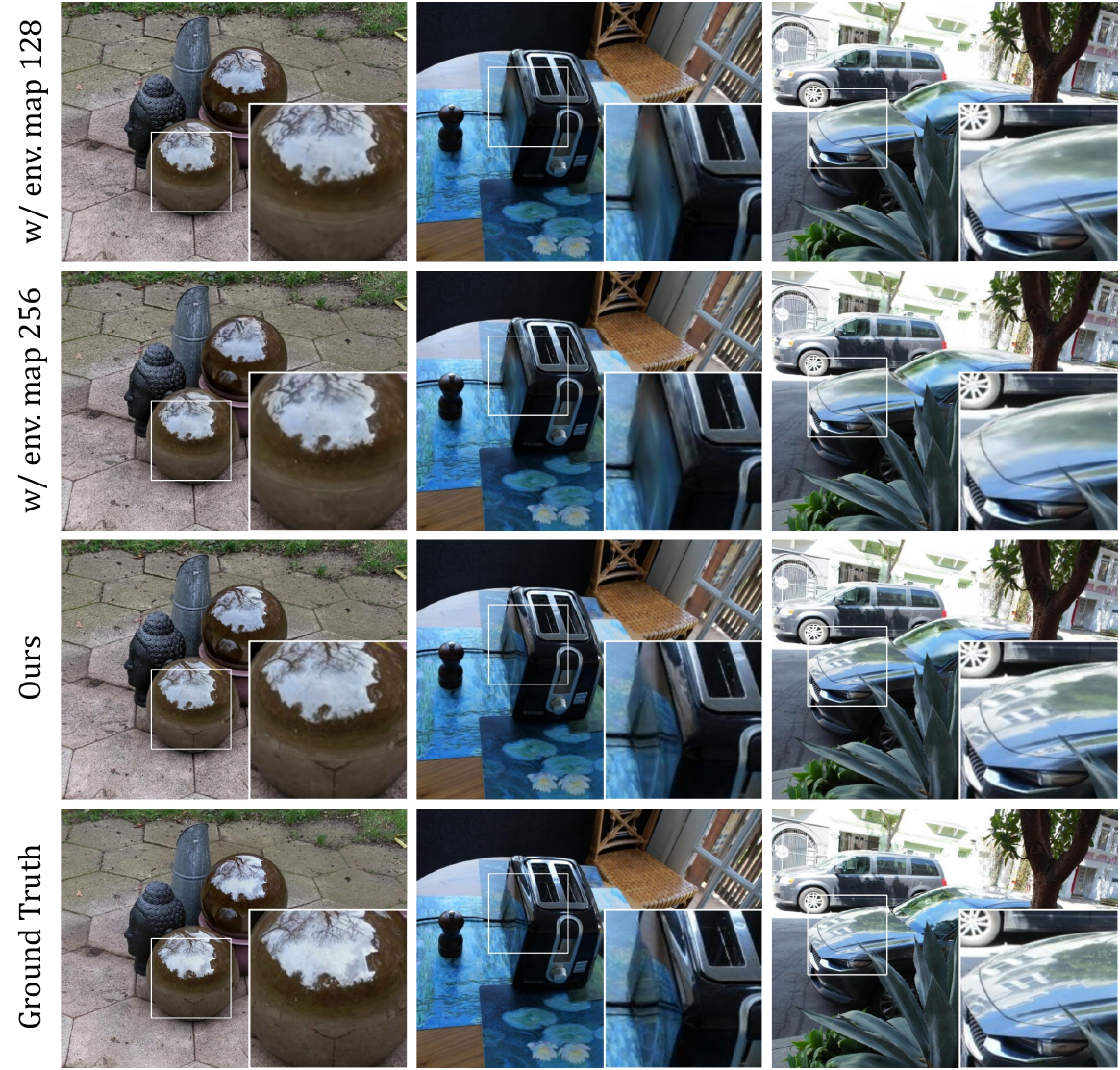}
    \caption{\textbf{Qualitative comparison between the environment map representation and our environment Gaussian representation}. Replacing the environment map with our environment Gaussian representation significantly improves the rendering quality, especially in capturing near-field reflections and high-frequency reflection details.}
    \label{fig:envgs_envmap}
\end{figure}

As described in Sec.~\ref{exp:ablation}, the environment representation plays a crucial role in capturing complex reflections. In Fig.~\ref{fig:envgs_envmap}, we provide additional qualitative comparisons between our \envgs{} representation and the environment map representation. The ``\textit{w/} env. map 128'' and ``\textit{w/} env. map 256'' variants replace our core \envgs{} representation with environment maps using six cubemaps at resolutions of 128 and 256, respectively. The results demonstrate that both environment map variants fail to capture the near-field reflections and tend to blur high-frequency reflection details, whereas our \envgs{} representation excels at capturing complex reflections with high fidelity.

\subsection{Speed Analysis}

We conduct additional speed ablations on the ``\textit{hatchback}'' from the NeRF-Casting Shiny Scenes~\cite{verbin2024nerf} with resolution 3504 $\times$ 2336 (which we downsample by a factor of 4, as done in all baselines and experiments). The results are listed in Tab.~\ref{tab:ablation_speed}.

\noindent\textbf{Differentiable Gaussian tracing.} As discussed in Sec.~\ref{method:tracing}, rendering the \envgs{} primitives with rasterization is impractical due to the uniqueness of each reflected ray. To validate this, we compare two alternative rendering strategies: (1) manually computing the ray-primitive intersections using PyTorch in a chunk-based manner (``\textit{w/} PyTorch''), and (2) rasterizing the \envgs{} primitives with a modified 3DGS~\cite{kerbl3Dgaussians} rasterizer using 1x1 tiles (``\textit{w/} 1x1-tile rasterizer''). All three methods, including our Gaussian tracer, apply the same volume rendering equation as in Eq.~\ref{eq:volume_rendering}. Quantitative results in Tab.~\ref{tab:ablation_speed} reveal that both alternative strategies take over an hour to render a single frame, whereas our Gaussian tracer achieves real-time rendering speeds, leveraging hardware-accelerated ray tracing.

\begin{table}[h]
    \setlength{\tabcolsep}{3pt}  
    \renewcommand{\arraystretch}{0.7}  
    \centering  
    \footnotesize  

    \begin{tabular}{lcccccc}
        \toprule
        &PSNR $\uparrow$ &SSIM $\uparrow$ &LPIPS $\downarrow$ &FPS $\uparrow$ \\

        \midrule
        \textit{w/} PyTorch             &-                &-               &-               &1/6157.613        \\
        \textit{w/} 1x1-tile rasterizer &-                &-               &-               &1/11902.431       \\

        \midrule
        \textit{w/} 50\% weight filting &27.137           &0.824           &0.192           &36.216            \\
        \textit{w/} 75\% weight filting &27.104           &0.823           &0.192           &41.824            \\
        \textit{w/} 80\% weight filting &27.033           &0.822           &0.193           &44.215            \\
        \textit{w/} 90\% weight filting &26.695           &0.816           &0.197           &\textbf{47.149}   \\

        \midrule
        Ours                            &\textbf{27.220}  &\textbf{0.838}  &\textbf{0.177}  &32.259            \\
        \bottomrule
    \end{tabular}
    \caption{\textbf{Runtime analysis of the proposed method on the \textit{hatchback} of NeRF-Casting Shiny Scenes~\cite{verbin2024nerf}.} Rasterization or PyTorch-based ray tracing is impractical for rendering the \envgs{} primitives. The acceleration techniques lead to minimal quality changes as shown by the cell.}
    \label{tab:ablation_speed}

  \end{table}

\noindent\textbf{Rendering speed analysis.} As mentioned in Sec.~\ref{method:representation}, the rendering of our method consists of two main rounds: rasterization of the \basegs{} and ray tracing of the \envgs{}, and the final color is the blending of the two. Based on the fact that only a small portion of the scene surface contains strong specular reflections, we can further accelerate the rendering process by only tracing rays with high blending weights, which are only made possible by our tracing-based renderer. We ablate the effectiveness and quality impact of this acceleration technique, results are shown in Tab.~\ref{tab:ablation_speed}.

\subsection{Environment Gaussian Design}

\noindent\textbf{The Necessity of using a separate environment Gaussian primitives.} To evaluate the decision to use separate Gaussian primitives for reflection modeling, we perform an experiment using a single set of Gaussian primitives for both reflection and base scene modeling. We first trace a camera ray to obtain the base color, normal and rendering weight, then trace a secondary ray to render the reflection color, ultimately combining these results using Eq. (\ref{eq:blendering}) to get the final color.
However, we found that the experiment consistently failed to converge due to unavoidable interference between suboptimal geometry during optimization and incorrectly hitting Gaussian primitives from erroneous reflection directions, leading to an unstable training process.

\section{Details of Environment Gaussian}
\label{sup:envgs}
We provide more details of our \basegs{} and \envgs{}. The SH coefficients of both \basegs{} and \envgs{} are set to two for the best results. The \envgs{} is jointly optimized with the \basegs{}, and \envgs{} constitutes around 15\% of the \basegs{}, of average 300k Gaussian primitives using 70MB after training.
For pruning, we follow the pruning method in the original 2DGS~\cite{huang20242d} and keep at most the top 630k \envgs{} primitives based on rendering weights.

\section{Details of Gradient Computation}
\label{sup:gradient}
To enable the joint optimization of \basegs{} and \envgs{} primitives, which is essential for accurate geometry recovery and reflection reconstruction (as demonstrated in Sec.~\ref{exp:ablation}), our Gaussian tracer must be fully differentiable. This requires computing gradients with respect to the input reflected ray origin, $\frac{\mathrm{d} \mathcal{L}}{\mathrm{d} \mathbf{o}}$, and direction, $\frac{\mathrm{d} \mathcal{L}}{\mathrm{d} \mathbf{d}}$. These gradients are backpropagated through the surface position $\mathbf{x}$ and normal $\mathbf{n}$, obtained during the first rasterization stage, to the \basegs{} parameters for joint optimization.

Consider an input ray with origin $\mathbf{o}$ and direction $\mathbf{d}$, and a intersected triangle primitive $i$ with vertices $\mathbf{v}_1, \mathbf{v}_2, \mathbf{v}_3$. During ray traversal, the OptiX kernel utilizes the GPU's RT core to determine the intersection depth $t_i$, which is then used to compute the interaction position as $\mathbf{x}_{i} = \mathbf{o} + t_i \mathbf{d}$. This position is subsequently transformed into the local tangent plane of the corresponding 2D Gaussian, yielding $\mathbf{u}_i$ via Eq.~\ref{eq:uv} for Gaussian value evaluation. Note that the ray-triangle intersection depth can be manually computed as:
\begin{equation}
    t_i = \frac{\mathbf{n}_i^\top (\mathbf{v}_1 - \mathbf{o})}{\mathbf{n}_i^\top \mathbf{d}},
\end{equation}
where $\mathbf{n}_i = (\mathbf{v}_2 - \mathbf{v}_1) \times (\mathbf{v}_3 - \mathbf{v}_1)$ is the normal direction of the triangle. Then, we can apply the chain rule to calculate the derivatives w.r.t. the ray origin and direction:
\begin{equation}
    \begin{aligned}
        \frac{\mathrm{d} \mathcal{L}}{\mathrm{d} \mathbf{o}} & = \frac{\mathrm{d} \mathcal{L}}{\mathrm{d} \mathbf{x}_i} \frac{\mathrm{d} \mathbf{x}_i}{\mathrm{d} \mathbf{o}}  + \frac{\mathrm{d} \mathcal{L}}{\mathrm{d} t} \frac{\mathrm{d} t}{\mathrm{d} \mathbf{o}} \\
         & = \frac{\mathrm{d} \mathcal{L}}{\mathrm{d} \mathbf{x}_i} + \frac{\mathrm{d} \mathcal{L}}{\mathrm{d} t} \cdot \frac{-\mathbf{n}_i}{\mathbf{n}_i^\top \mathbf{d}},
    \end{aligned}
\end{equation}
and
\begin{equation}
    \begin{aligned}
        \frac{\mathrm{d} \mathcal{L}}{\mathrm{d} \mathbf{d}} & = \frac{\mathrm{d} \mathcal{L}}{\mathrm{d} \mathbf{x}_i} \frac{\mathrm{d} t}{\mathrm{d} \mathbf{d}} + \frac{\mathrm{d} \mathcal{L}}{\mathrm{d} t} \frac{\mathrm{d} t}{\mathrm{d} \mathbf{d}} \\
        & = \frac{\mathrm{d} \mathcal{L}}{\mathrm{d} \mathbf{x}_i} \cdot t_i + \frac{\mathrm{d} \mathcal{L}}{\mathrm{d} t} \cdot \frac{-\mathbf{n}_i^\top (\mathbf{v}_1 - \mathbf{o})}{\mathbf{n}_i \cdot (\mathbf{n}_i^\top \mathbf{d})^2}.
    \end{aligned}
\end{equation}

This gradient flow enables the joint optimization of the reflection appearance of \envgs{} alongside the geometry and base appearance of \basegs{}, enhancing both geometry accuracy and reflection fidelity.

\begin{table*}[t!]
    \centering
    \small

        \begin{tabular}{llccccccc}
            \toprule
            \multirow{2.5}{*}{\bfseries PSNR $\uparrow$} & \multirow{2.5}{*}{\bfseries Methods} &\multicolumn{3}{c}{\bfseries Ref-Real~\cite{verbin2022refnerf}} &\multicolumn{4}{c}{\bfseries NeRF-Casting Shiny Scenes~\cite{verbin2024nerf}}  \\
            \cmidrule(lr){3-5} \cmidrule(lr){6-9}
                                     & &\textit{sedan} &\textit{toycar} &\textit{spheres} &\textit{compact} &\textit{grinder} &\textit{hatchback} &\textit{toaster} \\

            \midrule
            \multirow{4}{*}{\textit{Non real-time}}
            &Ref-NeRF*~\cite{verbin2022refnerf} &\cellthird25.390            &22.750            &21.120            &\cellsecond30.550 &\cellthird33.910            &25.210            &32.660            \\
            &UniSDF~\cite{wang2023unisdf}       &24.680            &\cellsecond24.150 &\cellsecond22.270 &29.720            &33.720            &\cellthird27.010            &\cellsecond32.900 \\
            &ZipNeRF~\cite{barron2023zipnerf}   &\cellsecond25.850 &\cellthird23.410            &\cellthird21.770            &\cellfirst31.100  &\cellfirst34.670  &\cellfirst27.780  &\cellfirst33.410	 \\
            &NeRF-Casting~\cite{verbin2024nerf} &\cellfirst26.770  &\cellfirst24.200  &\cellfirst23.040  &\cellthird29.730            &\cellsecond34.000 &\cellsecond27.490 &\cellthird32.870            \\

            \midrule
            \multirow{5}{*}{\textit{Real-time}}
            &3DGS~\cite{kerbl3Dgaussians}                  &\cellthird25.240            &\cellthird23.910            &\cellthird21.950            &\cellsecond28.945  &\cellsecond30.885 &\cellsecond26.201 &\cellsecond29.410 \\
            &2DGS~\cite{huang20242d}                       &25.065            &\cellsecond24.282 &\cellsecond22.064 &28.415            &\cellthird30.164            &25.893            &28.630            \\
            &GaussianShader~\cite{jiang2023gaussianshader} &24.081            &23.137            &21.408            &27.474            &26.572            &24.959            &26.641            \\
            &3DGS-DR~\cite{ye20243d}                       &\cellsecond25.445 &23.582            &21.539            &\cellthird28.692            &30.129            &\cellthird25.985            &\cellthird29.141            \\
            &Ours                                          &\cellfirst26.156  &\cellfirst24.746  &\cellfirst22.949  &\cellfirst29.608  &\cellfirst33.331  &\cellfirst27.220  &\cellfirst31.618  \\
            \bottomrule
            \addlinespace[8pt]

            \toprule            
            \multirow{2.5}{*}{\bfseries SSIM $\uparrow$} & \multirow{2.5}{*}{\bfseries Methods} &\multicolumn{3}{c}{\bfseries Ref-Real~\cite{verbin2022refnerf}} &\multicolumn{4}{c}{\bfseries NeRF-Casting Shiny Scenes~\cite{verbin2024nerf}}  \\
            \cmidrule(lr){3-5} \cmidrule(lr){6-9}
                                     & &\textit{sedan} &\textit{toycar} &\textit{spheres} &\textit{compact} &\textit{grinder} &\textit{hatchback} &\textit{toaster} \\

            \midrule
            \multirow{4}{*}{\textit{Non real-time}}
            &Ref-NeRF*~\cite{verbin2022refnerf} &\cellthird0.721            &0.612            &0.542            &\cellsecond0.907 &\cellthird0.880            &0.842            &0.932            \\
            &UniSDF~\cite{wang2023unisdf}       &0.700            &\cellsecond0.639 &\cellsecond0.567 &\cellthird0.895            &0.879            &\cellthird0.845            &\cellthird0.937            \\
            &ZipNeRF~\cite{barron2023zipnerf}   &\cellsecond0.733 &\cellthird0.626            &\cellthird0.545            &\cellfirst0.913  &\cellfirst0.887  &\cellfirst0.870  &\cellfirst0.944  \\
            &NeRF-Casting~\cite{verbin2024nerf} &\cellfirst0.739  &\cellfirst0.641  &\cellfirst0.597  &0.884            &\cellsecond0.882 &\cellsecond0.853 &\cellsecond0.938 \\

            \midrule
            \multirow{5}{*}{\textit{Real-time}}
            &3DGS~\cite{kerbl3Dgaussians}                  &\cellthird0.713            &\cellthird0.636            &\cellthird0.573             &\cellfirst0.877  &\cellsecond0.864 &\cellfirst0.838  &\cellsecond0.928 \\
            &2DGS~\cite{huang20242d}                       &0.704            &\cellsecond0.662 &\cellsecond0.595  &\cellthird0.857            &\cellthird0.854            &\cellsecond0.819            &\cellthird0.917            \\
            &GaussianShader~\cite{jiang2023gaussianshader} &0.668            &0.625            &\cellthird0.573             &0.851            &0.799            &0.805            &0.884            \\
            &3DGS-DR~\cite{ye20243d}                       &\cellsecond0.714 &0.635            &0.571             &\cellthird0.857            &0.849            &\cellthird0.813            &0.914            \\
            &Ours                                          &\cellfirst0.727  &\cellfirst0.667  &\cellfirst0.619  &\cellsecond0.871            &\cellfirst0.895  &\cellfirst0.838 &\cellfirst0.938  \\
            \bottomrule
            \addlinespace[8pt]

            \toprule
            \multirow{2.5}{*}{\bfseries LPIPS $\downarrow$} & \multirow{2.5}{*}{\bfseries Methods} &\multicolumn{3}{c}{\bfseries Ref-Real~\cite{verbin2022refnerf}} &\multicolumn{4}{c}{\bfseries NeRF-Casting Shiny Scenes~\cite{verbin2024nerf}}  \\
            \cmidrule(lr){3-5} \cmidrule(lr){6-9}
                                        & &\textit{sedan} &\textit{toycar} &\textit{spheres} &\textit{compact} &\textit{grinder} &\textit{hatchback} &\textit{toaster} \\

            \midrule
            \multirow{4}{*}{\textit{Non real-time}}
            &Ref-NeRF*~\cite{verbin2022refnerf} &\cellthird0.270            &0.257            &\cellthird0.257            &\cellsecond0.105 &0.123            &\cellthird0.156            &0.111            \\
            &UniSDF~\cite{wang2023unisdf}       &0.309            &\cellsecond0.245 &\cellsecond0.243 &\cellthird0.122            &\cellthird0.132            &0.160            &\cellthird0.107            \\
            &ZipNeRF~\cite{barron2023zipnerf}   &\cellsecond0.260 &\cellfirst0.243  &\cellfirst0.238  &\cellfirst0.096  &\cellfirst0.111  &\cellfirst0.130  &\cellfirst0.082  \\
            &NeRF-Casting~\cite{verbin2024nerf} &\cellfirst0.254  &\cellthird0.246            &\cellfirst0.238  &0.148            &\cellsecond0.114 &\cellsecond0.155 &\cellsecond0.096 \\

            \midrule
            \multirow{5}{*}{\textit{Real-time}}
            &3DGS~\cite{kerbl3Dgaussians}                  &\cellsecond0.301 &\cellsecond0.237 &\cellsecond0.248 &\cellfirst0.154  &\cellsecond0.181 &\cellsecond0.179  &\cellsecond0.123 \\
            &2DGS~\cite{huang20242d}                       &0.344            &\cellthird0.246            &0.254            &0.193            &\cellthird0.217            &\cellthird0.215            &0.147            \\
            &GaussianShader~\cite{jiang2023gaussianshader} &0.371            &0.293            &0.278            &\cellthird0.189            &0.289            &0.217            &0.169            \\
            &3DGS-DR~\cite{ye20243d}                       &\cellthird0.322            &0.249            &\cellthird0.251            &0.196            &0.219            &0.228            &\cellthird0.146            \\
            &Ours                                          &\cellfirst0.287  &\cellfirst0.208  &\cellfirst0.229  &\cellsecond0.159 &\cellfirst0.151  &\cellfirst0.177 &\cellfirst0.110  \\
            \bottomrule
        \end{tabular}

    \caption{\textbf{Ref-Real~\cite{verbin2022refnerf} and NeRF-Casting~\cite{verbin2024nerf} per-scene breakdowns.} All metrics are evaluated at the original resolution downsample by a factor of 4 as prior works ~\cite{verbin2024nerf}.}
    \label{tab:real_casting_break}
\end{table*}

\begin{table*}[t!]
    \centering
    \small

        \begin{tabular}{llcccccc}
            \toprule
            \multirow{2.5}{*}{\bfseries PSNR $\uparrow$} & \multirow{2.5}{*}{\bfseries Methods} &\multicolumn{6}{c}{\bfseries Shiny Blender Scenes~\cite{verbin2022refnerf}} \\
            \cmidrule(lr){3-8}
                                     & &\textit{ball} &\textit{car} &\textit{coffee} &\textit{helmet} &\textit{teapot} &\textit{toaster} \\

            \midrule
            \multirow{3}{*}{\textit{Non real-time}} 
            &Ref-NeRF~\cite{verbin2022refnerf}  &\cellfirst47.460  &\cellfirst30.820  &\cellfirst34.210  &\cellthird29.680            &\cellthird47.900            &\cellthird25.700	          \\
            &UniSDF~\cite{wang2023unisdf}       &\cellthird44.100            &\cellthird29.860            &\cellthird33.170            &\cellsecond38.840 &\cellsecond48.760 &\cellsecond26.180 \\
            &NeRF-Casting~\cite{verbin2024nerf} &\cellsecond45.460 &\cellsecond30.450 &\cellsecond33.180 &\cellfirst39.100  &\cellfirst49.980  &\cellfirst26.190  \\

            \midrule
            \multirow{6}{*}{\textit{Real-time}}
            &3DGS~\cite{kerbl3Dgaussians}                  &27.650            &27.260            &32.300            &28.220            &45.710            &20.990            \\
            &2DGS~\cite{huang20242d}                       &25.990            &26.730            &32.360            &27.300            &44.940            &20.272            \\
            &GaussianShader~\cite{jiang2023gaussianshader} &\cellthird29.081            &26.940            &31.147            &\cellthird28.883            &43.379            &\cellthird23.584            \\
            &3DGS-DR~\cite{ye20243d}                       &\cellfirst33.533  &\cellsecond30.236 &\cellfirst34.580  &\cellfirst31.518  &\cellfirst47.038  &\cellsecond26.823 \\
            &3iGS~\cite{tang20243igs}                      & 27.640	          &\cellthird27.510            &\cellthird32.580            &28.210            &\cellthird46.040            &22.690            \\
            &Ours                                          &\cellsecond32.567 &\cellfirst30.598  &\cellsecond34.312 &\cellsecond31.470 &\cellsecond46.582 &\cellfirst27.427  \\
            \bottomrule
            \addlinespace[8pt]

            \toprule
            \multirow{2.5}{*}{\bfseries SSIM $\uparrow$} & \multirow{2.5}{*}{\bfseries Methods} &\multicolumn{6}{c}{\bfseries Shiny Blender Scenes~\cite{verbin2022refnerf}} \\
            \cmidrule(lr){3-8}
                                     & &\textit{ball} &\textit{car} &\textit{coffee} &\textit{helmet} &\textit{teapot} &\textit{toaster} \\

            \midrule
            \multirow{3}{*}{\textit{Non real-time}}
            &Ref-NeRF~\cite{verbin2022refnerf}  &\cellfirst 0.995 &\cellsecond0.955 &\cellfirst0.974  &\cellthird0.958            &\cellsecond0.998 &\cellthird0.922            \\
            &UniSDF~\cite{wang2023unisdf}       &\cellthird0.993	          &\cellthird0.954            &\cellsecond0.973 &\cellfirst0.990  &\cellsecond0.998 &\cellsecond0.945 \\
            &NeRF-Casting~\cite{verbin2024nerf} &\cellsecond0.994 &\cellfirst0.964  &\cellsecond0.973 &\cellsecond0.988 &\cellfirst0.999  &\cellfirst0.950  \\

            \midrule
            \multirow{6}{*}{\textit{Real-time}}
            &3DGS~\cite{kerbl3Dgaussians}                  &0.937            &0.931            &0.972            &0.951            &\cellsecond0.996 &0.894            \\
            &2DGS~\cite{huang20242d}                       &0.935            &\cellthird0.932            &\cellthird0.973            &0.952            &\cellfirst0.997  &0.892            \\
            &GaussianShader~\cite{jiang2023gaussianshader} &\cellthird0.955            &0.930            &0.969            &\cellthird0.955            &\cellsecond0.996 &0.907            \\
            &3DGS-DR~\cite{ye20243d}                       &\cellfirst0.979  &\cellsecond0.957 &\cellfirst0.976  &\cellfirst0.971  &\cellfirst0.997  &\cellsecond0.943 \\
            &3iGS~\cite{tang20243igs}                      &0.938            &0.930            &\cellthird0.973            &0.951            &\cellfirst0.997  &\cellthird0.908            \\
            &Ours                                          &\cellsecond0.971 &\cellfirst0.958  &\cellsecond0.974 &\cellsecond0.968 &\cellfirst0.997  &\cellfirst0.945  \\
            \bottomrule
            \addlinespace[8pt]

            \toprule
            \multirow{2.5}{*}{\bfseries LPIPS $\downarrow$} & \multirow{2.5}{*}{\bfseries Methods} &\multicolumn{6}{c}{\bfseries Shiny Blender Scenes~\cite{verbin2022refnerf}} \\
            \cmidrule(lr){3-8}
                                        & &\textit{ball} &\textit{car} &\textit{coffee} &\textit{helmet} &\textit{teapot} &\textit{toaster} \\

            \midrule
            \multirow{3}{*}{\textit{Non real-time}}
            &Ref-NeRF~\cite{verbin2022refnerf}  &\cellthird0.059            &\cellsecond0.041 &\cellsecond0.078 &\cellthird0.075            &\cellsecond0.004 &\cellthird0.095            \\
            &UniSDF~\cite{wang2023unisdf}       &\cellfirst0.039  &\cellthird0.047            &\cellsecond0.078 &\cellsecond0.021 &\cellsecond0.004 &\cellfirst0.072  \\
            &NeRF-Casting~\cite{verbin2024nerf} &\cellsecond0.044 &\cellfirst0.033  &\cellfirst0.074  &\cellfirst0.018  &\cellfirst0.002  &\cellsecond0.073 \\

            \midrule
            \multirow{6}{*}{\textit{Real-time}}
            &3DGS~\cite{kerbl3Dgaussians}                  &0.162            &0.047            &\cellsecond0.079 &0.081            &\cellsecond0.008 &0.125            \\
            &2DGS~\cite{huang20242d}                       &0.155            &0.051            &\cellthird0.080            &0.080            &\cellsecond0.008 &0.126            \\
            &GaussianShader~\cite{jiang2023gaussianshader} &\cellthird0.145           &0.066            &0.085            &0.086            &\cellthird0.011            &0.105            \\
            &3DGS-DR~\cite{ye20243d}                       &\cellfirst0.104  &\cellsecond0.038 &\cellfirst0.076  &\cellfirst0.050  &\cellfirst0.006  &\cellsecond0.082 \\
            &3iGS~\cite{tang20243igs}                      &0.156            &\cellthird0.045           &\cellfirst0.076  &\cellthird0.073            &\cellfirst0.006  &\cellthird0.099            \\
            &Ours                                          &\cellsecond0.138 &\cellfirst0.037  &0.085            &\cellsecond0.052 &\cellfirst0.006  &\cellfirst0.077  \\
            \bottomrule
        \end{tabular}

    \caption{\textbf{Quantitative results on Shiny Blender Scenes~\cite{verbin2022refnerf}.}}
    \label{tab:shiny_blender}
\end{table*}

\begin{table*}[t!]
    \centering
    \small

        \begin{tabular}{llccccccccc}
            \toprule
            \multirow{2.5}{*}{\bfseries PSNR $\uparrow$} & \multirow{2.5}{*}{\bfseries Methods} &\multicolumn{9}{c}{\bfseries Mip-NeRF 360~\cite{barron2022mipnerf360}} \\
            \cmidrule(lr){3-11}
                                     & &\textit{bicycle} &\textit{bonsai} &\textit{counter} &\textit{flowers} &\textit{garden} &\textit{kitchen} &\textit{room} &\textit{stump} &\textit{treehill} \\

            \midrule
            \multirow{4}{*}{\textit{Non real-time}}
            &Ref-NeRF*~\cite{verbin2022refnerf} &\cellthird24.910           &32.290           &26.020           &21.630           &\cellthird27.450           &31.610            &\cellsecond31.680 &\cellthird25.910            &21.790            \\
            &UniSDF~\cite{wang2023unisdf}       &24.670            &\cellthird32.860            &\cellsecond29.260 &\cellsecond21.830 &\cellsecond27.460 &\cellthird31.730            &31.250            &\cellsecond26.390 &\cellsecond23.510 \\
            &ZipNeRF~\cite{barron2023zipnerf}   &\cellfirst25.800  &\cellfirst34.460  &\cellfirst29.380  &\cellfirst22.400  &\cellfirst28.200  &\cellfirst32.500  &\cellfirst32.650  &\cellfirst27.550  &\cellfirst23.890  \\
            &NeRF-Casting~\cite{verbin2024nerf} &\cellsecond24.920 &\cellsecond33.810 &\cellthird28.840            &\cellthird21.750            &27.310            &\cellsecond32.260 &\cellthird31.660            &25.640            &\cellthird23.220            \\

            \midrule
            \multirow{5}{*}{\textit{Real-time}}
            &3DGS~\cite{kerbl3Dgaussians}                  &\cellfirst25.250  &\cellfirst31.980  &\cellsecond28.700 &\cellsecond21.520 &\cellsecond27.410 &\cellthird30.320            &\cellthird30.630            &\cellfirst26.550  &\cellsecond22.490 \\
            &2DGS~\cite{huang20242d}                       &24.741            &\cellthird31.246            &\cellthird28.107            &\cellthird21.131            &26.723            &\cellsecond30.372 &\cellsecond30.679 &\cellsecond26.123 &\cellthird22.427            \\
            &GaussianShader~\cite{jiang2023gaussianshader} &23.103            &29.278            &26.639            &20.267            &26.290            &27.125            &24.098            &24.668            &20.552            \\
            &3DGS-DR~\cite{ye20243d}                       &\cellthird24.869            &31.232            &27.730            &21.116            &\cellthird27.142            &28.999            &30.068            &\cellthird25.473            &21.344            \\
            &Ours                                          &\cellsecond25.209 &\cellsecond31.946 &\cellfirst29.017  &\cellfirst21.551  &\cellfirst27.709  &\cellfirst31.660  &\cellfirst31.020  &25.423            &\cellfirst22.686  \\
            \bottomrule
            \addlinespace[8pt]

            \toprule
            \multirow{2.5}{*}{\bfseries SSIM $\uparrow$} & \multirow{2.5}{*}{\bfseries Methods} &\multicolumn{9}{c}{\bfseries Mip-NeRF 360~\cite{barron2022mipnerf360}} \\
            \cmidrule(lr){3-11}
                                     & &\textit{bicycle} &\textit{bonsai} &\textit{counter} &\textit{flowers} &\textit{garden} &\textit{kitchen} &\textit{room} &\textit{stump} &\textit{treehill} \\
    
            \midrule
            \multirow{4}{*}{\textit{Non real-time}} 
            &Ref-NeRF*~\cite{verbin2022refnerf} &0.723            &0.935            &0.875            &0.592            &\cellsecond0.845 &\cellthird0.922            &\cellsecond0.914 &0.731            &0.634            \\
            &UniSDF~\cite{wang2023unisdf}       &\cellthird0.737           &\cellthird0.939            &\cellsecond0.888 &\cellsecond0.606 &\cellthird0.844            &0.919            &\cellsecond0.914 &\cellsecond0.759 &\cellsecond0.670 \\
            &ZipNeRF~\cite{barron2023zipnerf}   &\cellfirst0.769  &\cellfirst0.949  &\cellfirst0.902  &\cellfirst0.642  &\cellfirst0.860  &\cellfirst0.928  &\cellfirst0.925  &\cellfirst0.800  &\cellfirst0.681  \\
            &NeRF-Casting~\cite{verbin2024nerf} &\cellsecond0.747 &\cellsecond0.945 &\cellthird0.887            &\cellthird0.605            &0.836            &\cellsecond0.924 &\cellthird0.911            &\cellthird0.749            &\cellthird0.653            \\

            \midrule
            \multirow{5}{*}{\textit{Real-time}}
            &3DGS~\cite{kerbl3Dgaussians}                  &\cellfirst0.771  &\cellfirst0.938  &\cellfirst0.905  &\cellfirst0.605  &\cellfirst0.868  &\cellsecond0.922 &\cellfirst0.914  &\cellfirst0.775  &\cellfirst0.638  \\
            &2DGS~\cite{huang20242d}                       &\cellthird0.734            &\cellthird0.931            &\cellthird0.893            &0.575            &0.844            &\cellthird0.917            &\cellthird0.907            &\cellsecond0.756 &\cellthird0.618            \\
            &GaussianShader~\cite{jiang2023gaussianshader} &0.700            &0.917            &0.875            &0.541            &0.842            &0.888            &0.839            &0.701            &0.579            \\
            &3DGS-DR~\cite{ye20243d}                       &\cellsecond0.740 &\cellsecond0.933 &0.889            &\cellthird0.578            &\cellthird0.852            &0.908            &0.904	         &\cellthird0.750            &0.607            \\
            &Ours                                          &\cellthird0.734           &\cellsecond0.933 &\cellsecond0.899 &\cellsecond0.589 &\cellsecond0.854 &\cellfirst0.923  &\cellsecond0.910 &0.726            &\cellsecond0.621 \\
            \bottomrule
            \addlinespace[8pt]

            \toprule
            \multirow{2.5}{*}{\bfseries LPIPS $\downarrow$} & \multirow{2.5}{*}{\bfseries Methods} &\multicolumn{9}{c}{\bfseries Mip-NeRF 360~\cite{barron2022mipnerf360}} \\
            \cmidrule(lr){3-11}
                                        & &\textit{bicycle} &\textit{bonsai} &\textit{counter} &\textit{flowers} &\textit{garden} &\textit{kitchen} &\textit{room} &\textit{stump} &\textit{treehill} \\

            \midrule
            \multirow{4}{*}{\textit{Non real-time}}
            &Ref-NeRF*~\cite{verbin2022refnerf} &0.256            &\cellthird0.182            &0.213            &\cellthird0.317            &\cellsecond0.132 &\cellthird0.121            &\cellsecond0.206 &0.261            &0.294            \\
            &UniSDF~\cite{wang2023unisdf}       &\cellthird0.243            &0.184            &\cellthird0.206            &0.320            &\cellthird0.136            &0.124            &\cellsecond0.206 &\cellsecond0.242 &\cellsecond0.265 \\
            &ZipNeRF~\cite{barron2023zipnerf}   &\cellfirst0.208  &\cellfirst0.173  &\cellfirst0.185  &\cellfirst0.273  &\cellfirst0.118  &\cellfirst0.116  &\cellfirst0.196  &\cellfirst0.193  &\cellfirst0.242  \\
            &NeRF-Casting~\cite{verbin2024nerf} &\cellsecond0.231 &\cellsecond0.176 &\cellsecond0.203 &\cellsecond0.312 &0.142            &\cellsecond0.118 &\cellthird0.216            &\cellthird0.244            &\cellthird0.273            \\

            \midrule
            \multirow{5}{*}{\textit{Real-time}}
            &3DGS~\cite{kerbl3Dgaussians}                  &\cellfirst0.205   &\cellsecond0.205 &\cellsecond0.204 &\cellfirst0.336  &\cellfirst0.103  &\cellsecond0.129 &\cellsecond0.220 &\cellfirst0.210  &\cellfirst0.317  \\
            &2DGS~\cite{huang20242d}                       &0.267             &\cellthird0.227            &\cellthird0.229            &0.374            &0.145            &\cellthird0.146            &\cellthird0.243            &\cellthird0.258            &\cellthird0.374            \\
            &GaussianShader~\cite{jiang2023gaussianshader} & 0.275            &0.242            &0.243            &0.380            &\cellthird0.131            &0.170            &0.307            &0.277            &0.394            \\
            &3DGS-DR~\cite{ye20243d}                       &\cellthird0.254            &0.230            &0.231            &\cellthird0.368            &0.135            &0.151            &0.247            &\cellsecond0.248 &0.375            \\
            &Ours                                          & \cellsecond0.233 &\cellfirst0.180  &\cellfirst0.194  &\cellsecond0.339 &\cellsecond0.112 &\cellfirst0.120  &\cellfirst0.207  &0.262            &\cellsecond0.347 \\
            \bottomrule
        \end{tabular}

    \caption{\textbf{Quantitative results on Mip-NeRF 360~\cite{barron2022mipnerf360}.} The results in ``Non Real-time'' are borrowed from NeRF-Casting~\cite{verbin2024nerf}, and Ref-NeRF* is an improved version of Ref-NeRF~\cite{verbin2022refnerf} that uses Zip-NeRF's~\cite{barron2023zipnerf} geometry model.}
    \label{tab:mipnerf360}
\end{table*}


\end{document}